\documentclass{article}

\PassOptionsToPackage{numbers}{natbib}
\usepackage[preprint]{neurips_2021}

\usepackage[utf8]{inputenc} 
\usepackage[T1]{fontenc}    
\usepackage{hyperref}       
\usepackage{url}            
\usepackage{booktabs}       
\usepackage{amsfonts}       
\usepackage{nicefrac}       
\usepackage{microtype}      
\usepackage{xcolor}         
\usepackage{graphicx}       
\usepackage{wrapfig}
\usepackage{amsmath}
\usepackage{caption}
\usepackage{subcaption}
\usepackage[export]{adjustbox}
\usepackage{multirow}

\title{Layer Folding: Neural Network Depth Reduction using Activation Linearization}

\author{%
  Amir Ben Dror \\
  Samsung Israel Research Center\\ 
  \texttt{amir.b@samsung.com} \\
  \And
  Niv Zehngut \\
  Samsung Israel Research Center\\
  \texttt{niv.z.b@samsung.com} \\
  \And
  Avraham Raviv \\
  Samsung Israel Research Center\\
  \texttt{avraham.r.b@partner.samsung.com} \\
  \And
  Evgeny Artyomov \\
  Samsung Israel Research Center\\
  \texttt{evgeny.a@samsung.com} \\
  \And
  Ran Vitek \\
  Samsung Israel Research Center\\
  \texttt{ran.vitek@samsung.com} \\
  \And
  Roy J Jevnisek \\
  Samsung Israel Research Center\\
  \texttt{roy.jewnisek@samsung.com} \\
}

\begin{document}

\maketitle

\begin{abstract}
Despite the increasing prevalence of deep neural networks, their applicability in resource-constrained devices is limited due to their computational load. While modern devices exhibit a high level of parallelism, real-time latency is still highly dependent on networks’ depth. Although recent works show that below a certain depth, the width of shallower networks must grow exponentially, we presume that neural networks typically exceed this minimal depth to accelerate convergence and incrementally increase accuracy. This motivates us to transform pre-trained deep networks that already exploit such advantages into shallower forms. We propose a method that learns whether non-linear activations can be removed, allowing to fold consecutive linear layers into one. We apply our method to networks pre-trained on CIFAR-10 and CIFAR-100 and find that they can all be transformed into shallower forms that share a similar depth. Finally, we use our method to provide more efficient alternatives to MobileNetV2 and EfficientNet-Lite architectures on the ImageNet classification task. 
\end{abstract}

\section{Introduction}

Multiple works have studied the relation between expressiveness and neural networks’ depth. 
Early works \citep{delalleau2011shallow, pascanu2013number, bianchini2014complexity} showed that some deep neural networks cannot be represented by shallower networks unless those networks are significantly wider. 
More specifically, \citet{telgarsky2016benefits} showed that there are networks of depth~$k^{2}$, for which any approximating network of depth~$k$ requires width of~$\mathcal{O}(\mathrm{e}^{k})$. 
Such depth separation \citep{daniely2017depth}, highlighting the exponential parameter growth of shallow networks compared to deeper ones representing the same function, has been further studied for other types of functions \citep{DBLP:conf/iclr/LiangS17, yarotsky2017error, safran2017depth, poggio2017and,petersen2018optimal,bolcskei2019optimal}. 
While these findings suggest that a certain depth is required to preserve performance on a given task, many architectures are typically deeper than that.
The role of the added layers can be viewed by the unrolled iterative estimation \citep{DBLP:conf/iclr/GreffSS17} — a group of successive layers iteratively refine their estimates of the same features instead of computing entirely new representations. 
Many works have even enforced iterative estimation by utilizing recurrent architectures instead of feed forward architectures for non-sequential input data \citep{trigeorgis2016mnemonic, van2020going}. 
In addition to potentially higher accuracy by iterative refined representations, and despite optimization barriers associated with increased depth \citep{bengio1994learning, glorot2010understanding}, deeper networks have also been shown favorable to training in some cases, acting as preconditioners that accelerate convergence \citep{arora2018optimization}. 
Hence, some layers in deep neural networks can be regarded as crucial for depth separation while others for refining representations and facilitating optimization, implying varying contribution to networks’ expressiveness. 
Such observation plays an important role when efficiency is additionally considered as an optimization objective. 
Particularly, while reducing a network’s depth below a certain level may markedly impact the network’s accuracy, reduction up to that level may exhibit lower and even no impact and yet considerably improve the network’s efficiency. 
We treat this level as the Effective Degree of Non-Linearity (EDNL), as the depth of feed forward networks is determined by the number of their non-linear layers. 
We argue that some networks wield more layers than their EDNL, suggesting that their expressive power can be maintained with wider-but-shallower networks. 
This has been supported by recent works \citep{perez2018deep, huh2021low} showing that deeper networks have a simplicity bias and encourage low-rank solutions. 
Shallower networks are particularly advantageous for hardware accelerators and GPUs that leverage intra-layer parallelism and suffer from inter-layer computational overhead. 

We propose to learn which activations can be removed without incurring a significant accuracy degradation. 
This allows us to merge adjacent linear layers, and in turn, transform deep networks into shallow ones. 
Over-parameterization in feed forward networks does not solely relate to their depth but may also evident in layers’ width. 
Pruning methods \citep{lecun1990optimal, hassibi1993second, han2015learning} are commonly used to optimize networks’ width, with the ability to prune an entire layer. 
Similarly to pruning methods, we focus on optimizing a pre-trained network. 
This stems from an optimization perspective; first, the fine-tuning process involved in such optimization requires far less computational resources than training the refined architecture from scratch. 
Second, gradient-based optimization of the shallower network may better proceed from the original deeper network. 
This allows the network to leverage the rich representations and local minimum obtained by the deeper network form and distill it to its shallow form. 
Recent pruning methods \citep{zhuang2020neuron, tiwari2021chipnet} attenuate neurons during a fine-tuning phase, gradually reducing the network’s size while allowing it to compensate. 
However, when applied to layer pruning, these methods force the network to gradually adopt new intermediate representations. 
The deep and shallow forms of a network may reside in local minima, for which traversing from one another may be challenging by gradient descent. 
In contrast, our optimization method maintains the intermediate representations of the original deep network during the fine-tuning phase, followed by a functionality-preserving transformation. 
Recently, \citet{ding2021repvgg} suggested a similar decoupling between training-time architecture and inference-time architecture via structural re-parametrization to leverage parallel connections during training. 

In line with other optimization methods, we focus on Convolutional Neural Networks (CNNs) for their prevalence in compute-intensive vision applications. 
We learn to remove non-linear activations between consecutive convolution layers, allowing their functionality-preserving merge. 
For layers with spatial kernels of size $k \times k$, this result in a larger $(2k-1) \times (2k-1)$ kernel. 
We show how such a transformation, in spite of the added FLOPs, may reduce latency on different hardware devices. 
Interestingly, this comes in reverse of the common preference of multiple smaller kernels (e.g., $3 \times 3$) than fewer larger ones (e.g., $5 \times 5$) which originated in \citep{DBLP:journals/corr/SimonyanZ14a}. 
For some cases, our approach can be regarded as an inverse of network factorization methods that utilize tensor decompositions such as CP-decomposition and Tucker decomposition \citep{DBLP:journals/corr/KimPYCYS15, DBLP:conf/nips/NovikovPOV15, DBLP:journals/corr/TaiXWE15, DBLP:journals/corr/LebedevGROL14}. 
Indeed, merging consecutive layers of depth-wise separable convolutions as in MobileNetV1 \citep{howard2017mobilenets} may lead to far more computations and negatively impact latency. 
In contrast, for other architectures that incorporate bottlenecks, merging layers can in fact reduce FLOPs, and latency furthermore. 
For example, merging layers of a MobileNetV2 inverted bottleneck \citep{sandler2018mobilenetv2} into a single convolution layer may cut FLOPs in half. 
We experiment on this architecture to show such potential gain. 
The fact that many recent works \citep{wu2019fbnet, stamoulis2019single, howard2019searching, wan2020fbnetv2, nayman2021hardcore} have relied on neural architectures composed of inverted bottlenecks to achieve prominent performance over the ImageNet classification task \citep{ILSVRC15} highlights the attractiveness of our method. 

While most pruning methods may prune entire layers indirectly, we attend such optimization explicitly. 
Specifically, pruning methods penalize weights whereas our method penalizes activations. 
This implies that our method can be used jointly with other pruning methods, improving efficiency even further. 
In addition, our method result in merged weight kernels which are necessarily low-rank. 
For instance, a $5 \times 5$ weight kernel resulting from merging two $3 \times 3$ kernels is bounded to a dimension of $18$ rather than $25$. 
Further optimization may leverage this redundancy to reduce computations even further. 
For example, the aforementioned $5 \times 5$ kernel may be more easily distilled into a smaller $3 \times 3$ kernel compared to an arbitrary $5 \times 5$ kernel. 

Our contributions are as follows:
\begin{enumerate}
\item We propose \emph{Layer Folding}, a novel method to reduce the depth of a neural network and fold consecutive linear layers by removing the non-linear activations that separate them. We show how our method facilitates optimization by maintaining the intermediate representations of the original depth. 
\item We introduce EDNL, which provides information on the required depth for networks to maintain their functionality. We reveal the EDNL of ReLU-activated CNNs and show that it is closely related to the complexity of their task rather than their original depth. 
\item We apply our method on efficient mobile networks over the ImageNet classification task \citep{ILSVRC15} and improve their latency even further without a significant impact on their accuracy. 
\end{enumerate}

\section{Related Work}

\paragraph{Pruning.}
Similar to pruning, we modify the existing architecture of a pre-trained network in order to improve its efficiency while maintaining its accuracy.  
Pruning methods remove selected neurons – nodes, filters or layers – according to pruning criteria. 
Works such as \citep{lecun1990optimal, hassibi1993second, DBLP:conf/iclr/MolchanovTKAK17} rely on Taylor expansions provided by the Hessian matrix of the loss function. 
Other works \citep{han2015learning, han2016eie, wang2020neural} prune neurons according to their magnitude. 
Different importance assignment methods have also been proposed, such as score propagation \citep{yu2018nisp} or statistics drawn from adjacent layers \citep{luo2018thinet}. 
Resembling our focus on depth reduction, dedicated layer pruning methods have also been proposed. 
\citet{jordao2020discriminative} considered the discriminative power of feature maps as layer scoring which is then used as scoring mechanism. 
\citet{chen2018shallowing} proposed to estimate whether layers can be replaced with linear layers, sharing a similar motivation to ours. 
In contrast, they optimize layers independently while we train a network as a whole in an end-to-end manner, allowing layers to compensate for those which are removed. 
\citet{neill2020compressing} proposed to remove layers based on layer similarity. 
They also empirically showed that there is a bound on the amount of compression that can be achieved before an exponential degradation in performance.
We draw a different conclusion and show that such a bound originates from a network's depth rather than its size (i.e., number of parameters).
Our work is in line with pruning methods that perform fine-tuning with additional loss which encourage more efficient networks, such as \citep{xu2020layer}. 
However, while such methods require the network to learn new intermediate representations, we maintain the rich representation space of the original network. 
Moreover, according to \citep{DBLP:conf/mlsys/BlalockOFG20}, pruning entails masking or removing weight elements of a given model, producing a smaller one. 
Our work does not fall under this definition, as we introduce new weight elements, and may even increase the model’s size. 

\paragraph{Neural Architecture Search (NAS).} 
As high-capacity and high-performing neural networks became feasible to train, designing efficient architectures has gained high interest. 
Early works have suggested various design principles, such as adopting multiple smaller spatial kernels instead of larger ones \citep{DBLP:journals/corr/SimonyanZ14a}, leveraging residual connections \citep{he2016deep}, decomposing weight matrices \citep{howard2017mobilenets} and utilizing bottlenecks \citep{sandler2018mobilenetv2}. 
Current top-performing architectures are found using NAS. 
Since the search space spanned by possible architectures is intractable, existing methods use reinforcement learning \citep{zoph2016neural}, genetic algorithms \citep{real2019regularized}, differentiable search \citep{liu2018darts} and other methods \citep{DBLP:conf/nips/KandasamyNSPX18, liu2018progressive, luo2018neural} to traverse it. 
Similarly to our method, differentiable methods learn architectural paths that allows the removal of entire layers, and even compensate it with added width to the preceding ones. 
For example, considering a reference architecture, all the following methods \citep{wu2019fbnet, stamoulis2019single, howard2019searching, wan2020fbnetv2, nayman2021hardcore} may learn to remove a convolution layer while enlarging the kernel of the preceding layer. 
However, NAS methods in general and these methods in particular require costly computational resources for both training a super-network from scratch and covering multiple search space dimensions. 
Our method leverages a pre-trained network and focuses on a single search dimension: depth. 
Both allow an expedited training time and facilitated convergence. 

\paragraph{Activation removal.}
Our method uniquely combines activation removal with consecutive linear layer folding. 
Nevertheless, activation removal alone has been used for other purposes. 
\citet{he2015delving} proposed PReLU, generalizing the ReLU activation function. 
\citet{ma2020activate} extended this idea to arbitrary activation functions. 
These methods allow learning whether activations should be shifted towards identity. 
However, their method learns the extent of such shift with accuracy optimization in mind. 
In contrast, we learn a binary decision – keeping the original activation or replacing it with an identity – with efficiency optimization in mind. 
Since our objective comprises depth minimization on top of the original task loss, our method converges to different solutions. 
Activation removal has gained further attention with the resurgence of Private Inference (PI). 
PI performs inference on encrypted data, where latency is hindered mostly by non-linear activations such as ReLU. 
\citet{NEURIPS2020_c519d47c} performed NAS to optimize the placement of skip connections and considered pruning ReLU activations that follow them. 
\citet{DBLP:journals/corr/abs-2103-01396} proposed to measure ReLU criticality by evaluating a model’s performance with ReLUs of entire stages or alternating layers being removed. 
While we leave it to future work, our work can also be used to accelerate PI. 

\section{Method}
\label{method_section}
We present a method which allows to reduce the number of non-linear activations in a neural network. This effectively enables to merge adjacent linear layers into a single one. We call this process \emph{Layer Folding}.
Given a non-linear activation function $\sigma$, we define the parametric activation $\sigma_\alpha ()$ to be the linear combination of $\sigma$ and the identity function:

\begin{equation}
	\sigma_\alpha (x) = \alpha x + (1-\alpha) \sigma(x), \quad 0 \leq \alpha \leq 1
	\label{eq:parameteric_act}
\end{equation}

where $\alpha$ is a trainable parameter which provides an interpolation between $\sigma$ and the identity function. 
When $\sigma=ReLU$, $\sigma_\alpha ()$ is the common PReLU \citep{he2015delving} activation. 
Given a trained neural network $\mathcal{F}$, we construct a network $\mathcal{F}_{\alpha}$ by transforming its activations into their corresponding parametric activations initialized with $\alpha=0$. 
This ensures that $\mathcal{F}_{\alpha}$ maintains the same functionality of $\mathcal{F}$. 

We perform a fine-tuning stage in which we optimize $\mathcal{F}_{\alpha}$ with respect to both the original task loss $\mathcal{L}_t$ and an auxiliary loss $\mathcal{L}_c$ that penalizes smaller $\alpha$ values, encouraging them to become $1$. 
We consider a general form of $\mathcal{L}_c$:

\begin{equation}
    \mathcal{L}_c = \sum_{l \in L} c_l \, h(\alpha_l)
    \label{eq:generic_alpha_loss}
\end{equation}

where $\alpha_l$ corresponds to the $l$th activation and $h(\alpha)$ is a monotonically decreasing function for $0 \leq \alpha \leq 1$. 
$ \left\{ c_l \right\}_{l \in L}$ weigh the contribution of each layer to $\mathcal{L}_c$. 
These are used to depict a varying potential value for folding different layers and can be set, for example, according to a measured latency on a target device. 
When we simply want to encourage a shallow network we set $c_l = 1$, $l=1:L$. 

While many forms of $h$ can be applied, we provide a simple suggestion for $h$ such that $\mathcal{L}_c$ becomes:

\begin{equation}
    \mathcal{L}_c = \sum_{l \in L} c_l \, (1-\alpha_l^p)
    \label{eq:specific_alpha_loss}
\end{equation}

We select this form for the following reasons: after training, a layer can be folded with its subsequent one only if its corresponding $\alpha$ is sufficiently close to $1$. 
In particular, we are sensitive to small changes in $\alpha_l$ near $1$, since the farther $\sigma_\alpha$ deviates from identity the larger the error incurred from Layer Folding. 
Yet, we are indifferent to small changes in $\alpha$ near $0$ since the matching layer cannot be folded anyway. We would like $\mathcal{L}_c$ to represent these ideas. 
$p>1$ is a hyperparameter controlling the flatness of the loss surface around $\alpha_l=0$ and the strength in which larger $\alpha_l$ values are pushed to $1$. 
Our final loss function is:

%
%

\begin{equation}
    \mathcal{L} = \mathcal{L}_t + \lambda_c \, \mathcal{L}_c
	\label{eq:total_loss}
\end{equation}

where $\lambda_c$ is a hyperparameter that balances between the task loss and the amount of layers that will be folded. 

We define the folding of two adjacent linear layers that reside between the feature maps $\left\{X,Y,Z\right\}$, $g_1:X \rightarrow Y$ and $g_2:Y \rightarrow Z$, as their composite function, i.e., $g_1 \circ g_2:X \rightarrow Z$. 
We provide two examples for fully connected layers and convolution layers while omitting the linear bias addition and batch normalization operations for brevity. 
For $g_1(\mathbf{x})= \mathbf{W}_1 \mathbf{x}$, $g_2(\mathbf{y})=\mathbf{W}_2 \mathbf{y}$ fully connected layers where $\mathbf{x} \in \mathbb{R}^{d_X}$, $\mathbf{y} \in \mathbb{R}^{d_Y}$, $\mathbf{z} \in \mathbb{R}^{d_Z}$, $\mathbf{W}_1 \in \mathbb{R}^{d_Y \times d_X}$, $\mathbf{W}_2 \in \mathbb{R}^{d_Z \times d_Y}$, their folding is given by:

\begin{equation}
	g_{fold}(\mathbf{x})=\mathbf{W}_2 \mathbf{W}_1 \mathbf{x}
	\label{eq:fc_fold}
\end{equation}

For $g_1(\mathbf{x})= \left\{ \sum_{i=1}^{c_X} \mathbf{W}_1^{i,j} \ast \mathbf{x}_i \right\}_{j=1}^{c_Y}$, $g_2(\mathbf{y})= \left\{ \sum_{j=1}^{c_Y} \mathbf{W}_2^{j,m} \ast \mathbf{y}_j \right\}_{m=1}^{c_Z}$ convolution layers where $\mathbf{x} \in \mathbb{R}^{h \times w \times c_X}$, $\mathbf{y} \in \mathbb{R}^{h \times w \times c_Y}$, $\mathbf{z} \in \mathbb{R}^{h \times w \times c_Z}$, 
$\mathbf{W}_1 \in \mathbb{R}^{k \times k \times c_Y \times c_X}$, $\mathbf{W}_2 \in \mathbb{R}^{k \times k \times c_Z \times c_Y}$, their folding is given by:

\begin{equation}
	g_{fold}(x)= \left\{ \sum_{i=1}^{c_X} \left( \sum_{j=1}^{c_Y} W_1^{i,j} \ast W_2^{j,m} \right) \ast x_i \right\}_{m=1}^{c_Z}
	\label{eq:conv_fold}
\end{equation}

\begin{figure}
    \centering
    \includegraphics[width=1.0\textwidth]{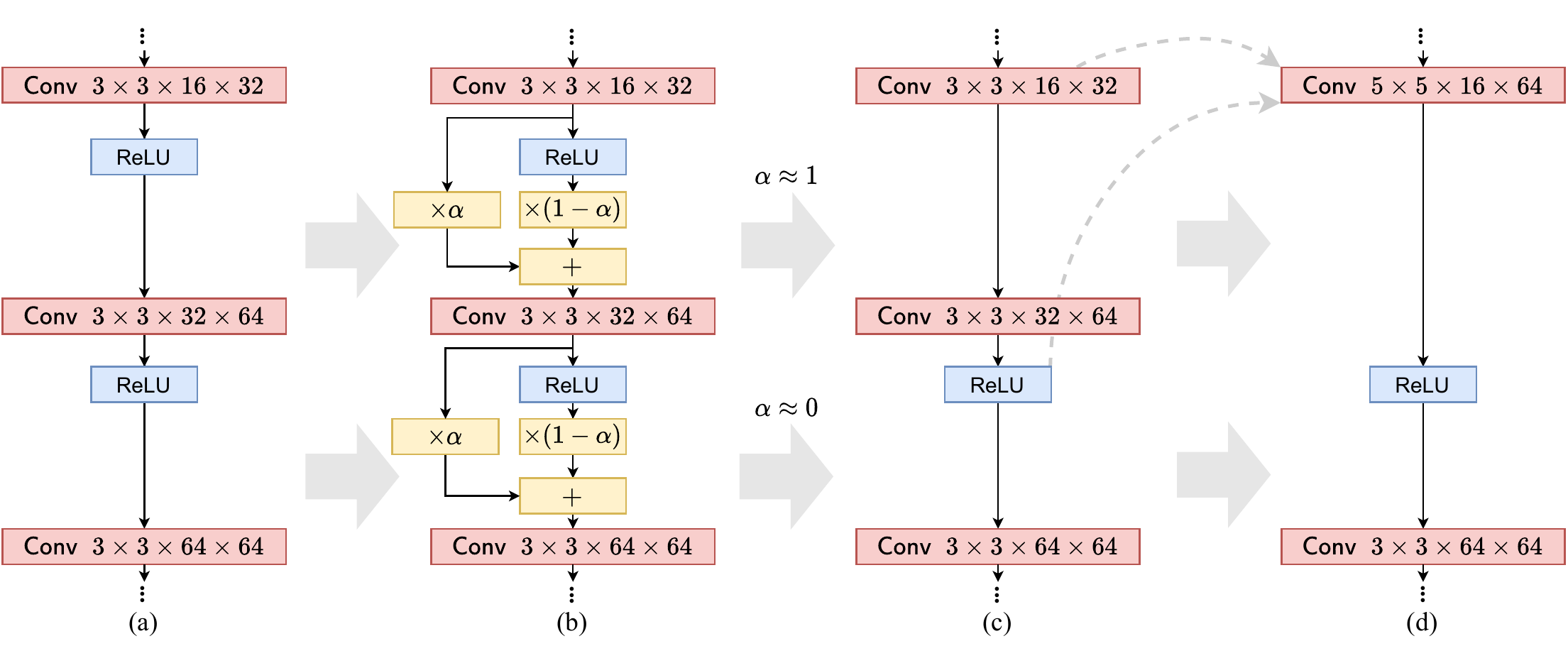}
    \caption{Illustration of our method. We replace  activations of a given network $\mathcal{F}$ (a) with parametric activations according to Equation \eqref{eq:parameteric_act}, resulting in $\mathcal{F}_{\alpha}$ (b). We fine-tune $\mathcal{F}_{\alpha}$ with the loss provided in Equation \eqref{eq:total_loss}. When training converges, we remove activations whose $\alpha\approx1$ (c). We fold consecutive linear operations (e.g., using Equation \eqref{eq:fc_fold} and Equation \eqref{eq:conv_fold}), resulting in a shallower network $\mathcal{F}_{fold}$ (d). We then fine-tune $\mathcal{F}_{fold}$.}
	\label{LF_illustration}    
\end{figure}

Our method comprises 2 phases: \emph{pre-folding} and \emph{post-folding}. 
In the pre-folding phase we fine-tune $\mathcal{F}_{\alpha}$ with the loss defined in~\eqref{eq:total_loss}.
When training converges we remove activations whose $\alpha$s exceed a threshold $\tau$ and fold the corresponding adjacent layers, resulting in a shallower network $\mathcal{F}_{fold}$. 
In the post-folding phase, we fine-tune $\mathcal{F}_{fold}$ once more for two main reasons. 
First, the underlying function of $\mathcal{F}_{fold}$ may yet deviate from $\mathcal{F}_{\alpha}$ due to various layers’ attributes such as padding, resulting in a small accuracy decrease. 
Post-folding fine-tuning allows the network to recover from it. 
Second, in some cases, $\mathcal{F}_{fold}$ result in a larger number of weights. 
Further training of $\mathcal{F}_{fold}$ may leverage the added capacity and increase accuracy. 
Our method is illustrated in Figure \ref{LF_illustration}. 

\section{Effective Degree of Non-Linearity Evaluation}
\label{EDNL_section}


In this section we use Layer Folding to evaluate the EDNL of several neural networks over image classification tasks. 
The success of neural networks is commonly attributed to their large capacity and non-linear nature. 
The particular contribution of networks’ size in the training phase motivates the pursuit for more efficient networks to be used for inference, and is often carried out by pruning and NAS methods. 
The contribution of the non-linearity of neural networks, however, is seldom assessed. 
Many works have shown that for various cases, neural networks must employ a certain depth unless exponentially increasing their width \citep{telgarsky2016benefits, DBLP:conf/iclr/LiangS17, yarotsky2017error, safran2017depth, poggio2017and,petersen2018optimal,bolcskei2019optimal}. 
Since depth, similarly to model's size, may be stretched to aid the training phase \citep{arora2018optimization}, networks’ depth can also be reduced for efficient inference. 
Unlike networks' size, their depth accounts for a complexity measure that we regard as the EDNL. 
Moreover, we expect the EDNL to be associated with a function that yields a certain accuracy over a certain task, and as such, that similarly performing networks will share an EDNL. 

In order to show that a network possesses an EDNL, we show that its accuracy is roughly maintained down to a certain depth and drops below it. 
Particularly, we ensure that this holds true even when the network’s size increases as its depth grows smaller. 
We further show that such depth knee-point is shared for different networks over a particular task. 



We perform our experiments on MNIST \citep{lecun1998gradient}, CIFAR-10 and CIFAR-100 \citep{krizhevsky2009learning} image classification tasks. 
For MNIST, we train fully-connected networks with depth $L \in [2:10]$, ReLU activation and width $d=256$ for all layers. 
We denote these networks by \emph{FC-$L$}.
For CIFAR-10 and CIFAR-100, we consider the commonly used ResNet and VGG architectures \citep{he2016deep, DBLP:journals/corr/SimonyanZ14a}. 
We use pre-trained ResNet models with depth $L \in \left\{ 20,32,44,56 \right\}$ and VGG models with depth $L \in \left\{ 16,19 \right\}$ \citep{chenyaofo2019}. 
For each of these networks, we apply Layer Folding with $c_l=1$, $l=1:L$, $p=2$, $\tau=0.9$ while varying $\lambda_c$ to obtain shallower networks of varying depth. 

\setlength{\columnsep}{0.05\textwidth}%
\begin{wrapfigure}[17]{r}{0.45\textwidth}
	\centering
	\vspace{-\intextsep}
    \includegraphics[width=0.45\textwidth, right]{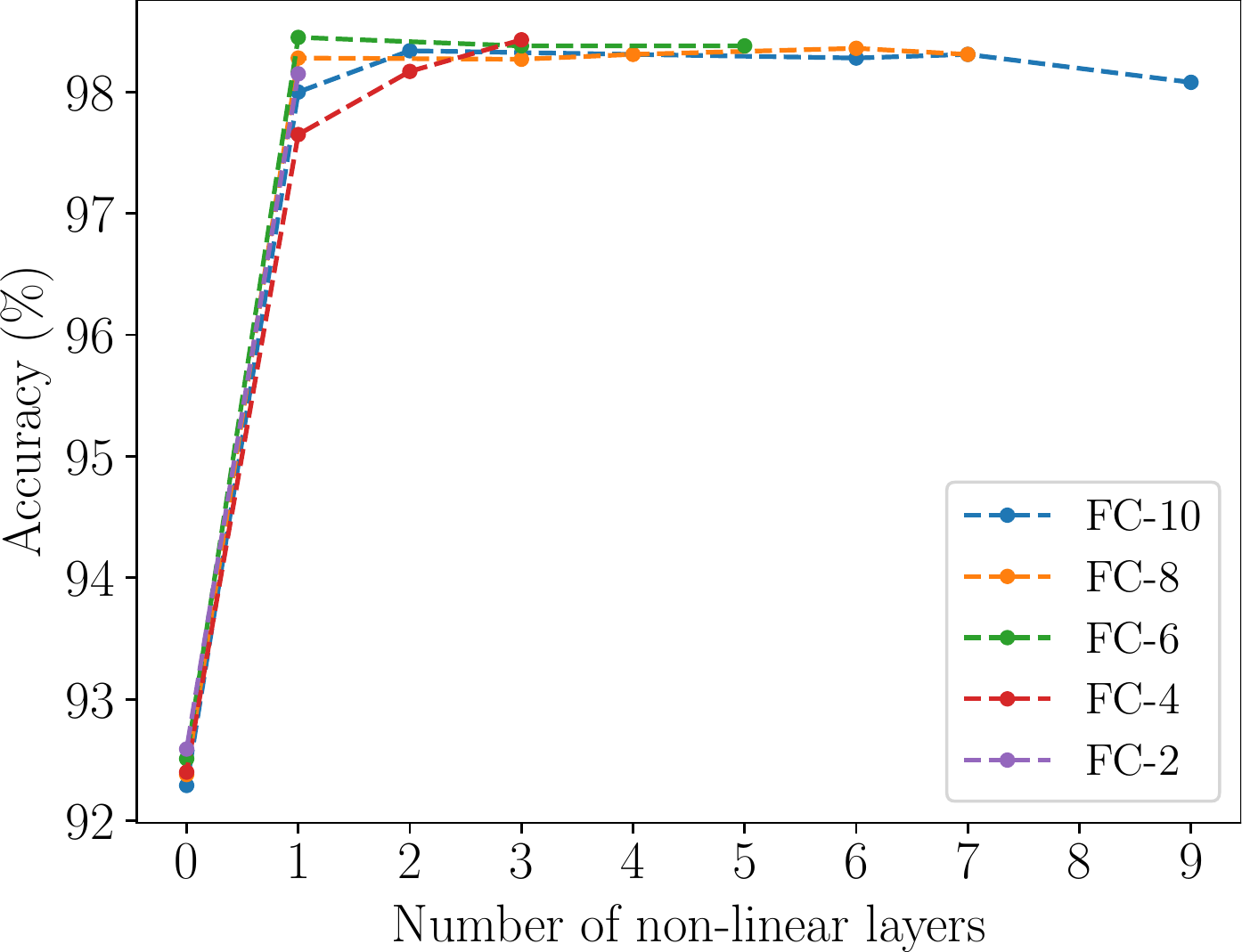}
    \caption{Layer Folding applied on fully-connected networks with depth $L \in [2:10]$. For each network, we gradually remove non-linear activations.}
	\label{1_MNIST_EDNL}    
\end{wrapfigure}

\begin{figure}
	\centering
	\begin{minipage}{0.49\linewidth}
	    \includegraphics[width=0.99\textwidth, left]{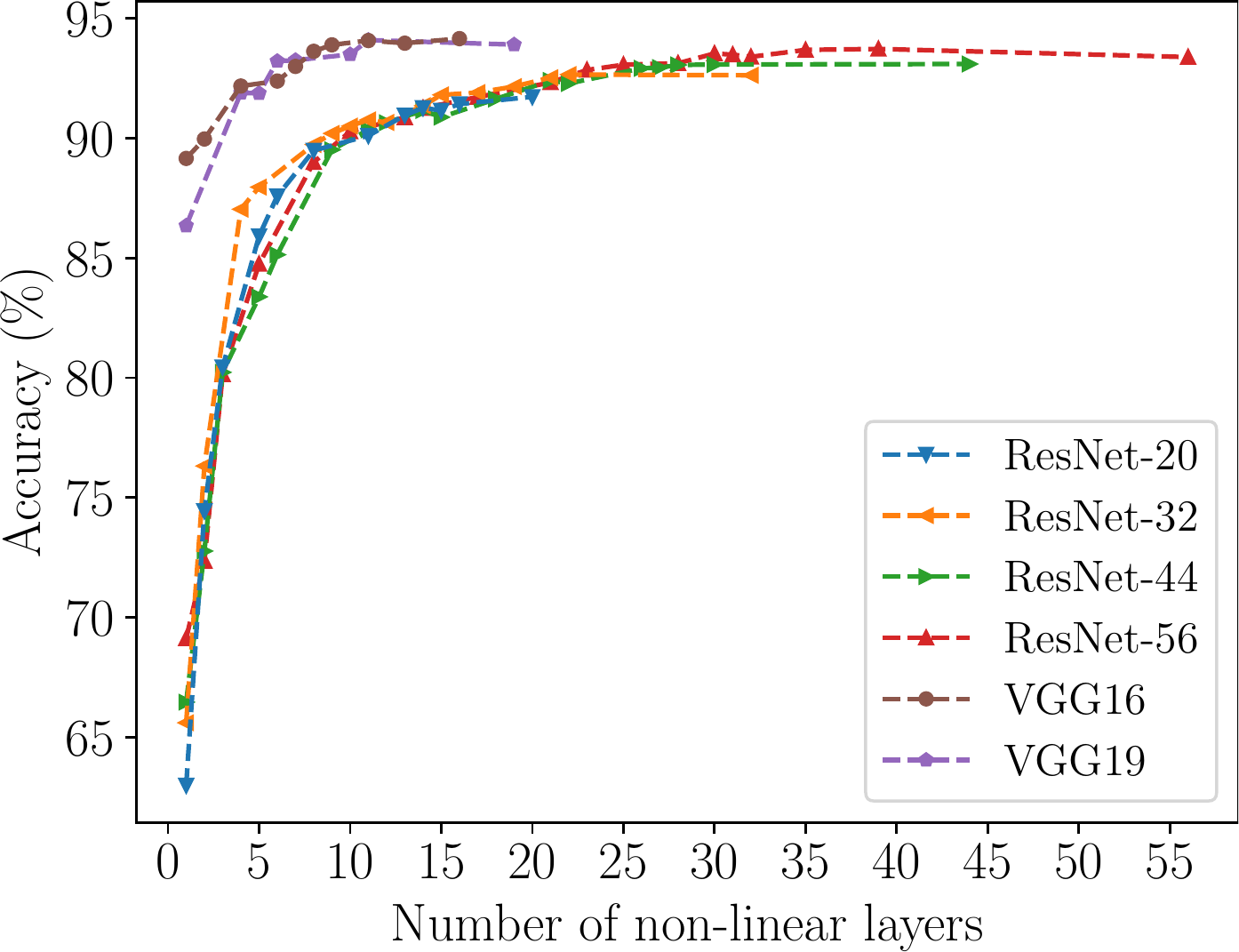}
		\label{2a_CIFAR10_ResNet_EDNL}
	\end{minipage}
	\;
	\begin{minipage}{0.49\linewidth}
	    \includegraphics[width=0.99\textwidth, right]{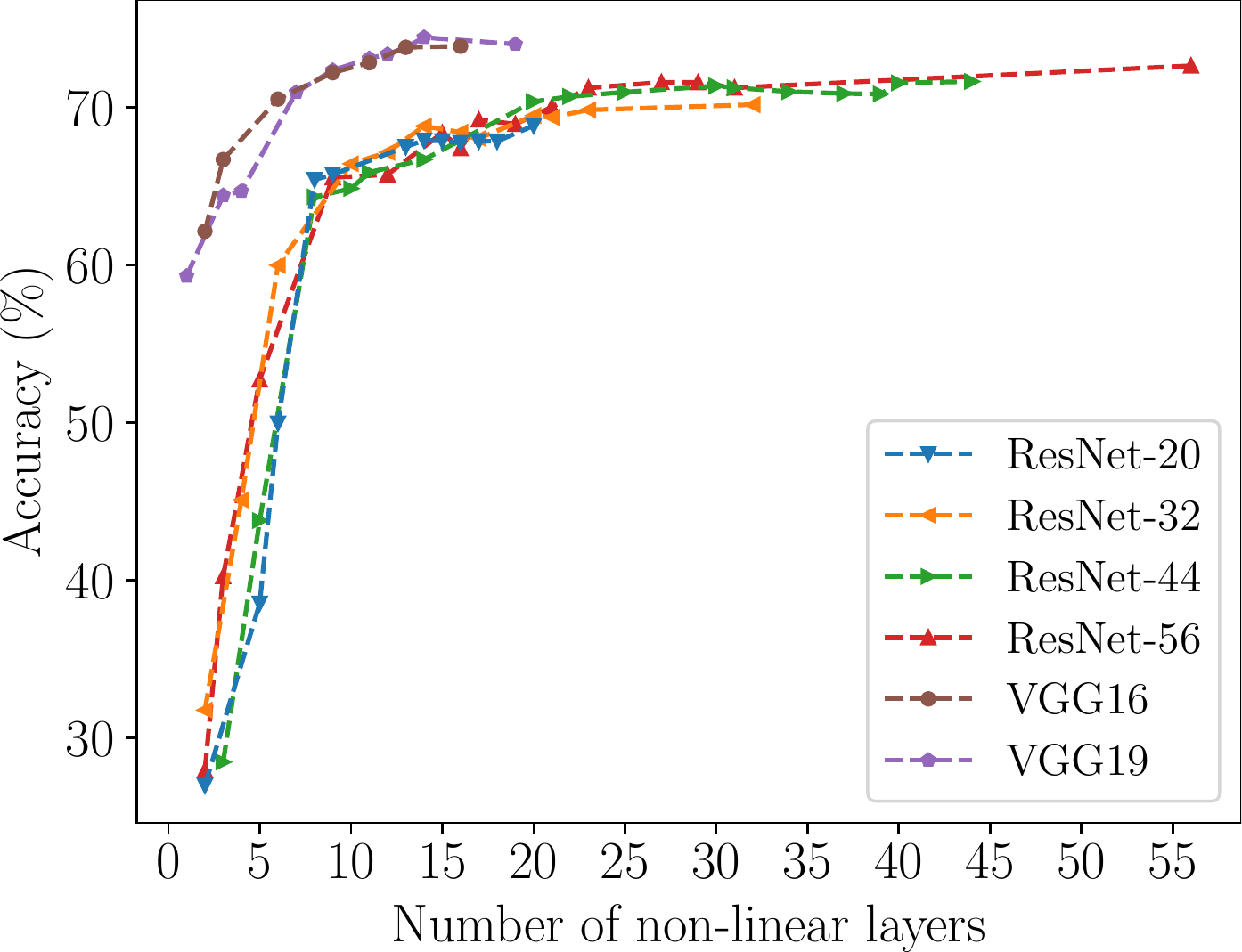}
		\label{2b_CIFAR100_ResNet_EDNL}
    \end{minipage}	
    \caption{Layer Folding applied on ResNet and VGG architectures on CIFAR-10 (left) and CIFAR-100 (right). For each network, we gradually remove non-linear layers. For CIFAR-10 (CIFAR-100), an accuracy drop is noticeable for number of non-linear layers below 8 (10).}
	\label{2_CIFAR_ResNet_EDNL}
\end{figure}

We first apply only pre-folding in order to evaluate the EDNL without increasing the networks' size. 
We count the number of remaining non-linear layers, denoted by $\hat{L}$. 
As shown in Figure \ref{1_MNIST_EDNL}, an EDNL is clearly visible for FC-$L$ networks, as they all retain their accuracy down to $\hat{L}=2$ which is then degraded for $\hat{L}=1$.
Figure \ref{2_CIFAR_ResNet_EDNL} shows our results for CIFAR-10 and CIFAR-100. 
For CIFAR-10, all ResNet models exhibit a small accuracy drop down to a depth of $\hat{L}=8$ and a large one below it. 
A similar phenomenon is observed for CIFAR-100. As expected, the classification task with the added classes exhibits a slightly larger EDNL. 
The gap between these EDNLs may indicate the general complexity increase for classifying $100$ classes instead of $10$ or point to a more complex feature extraction required for some of the added classes.

Figure \ref{2_CIFAR_ResNet_EDNL} also shows our results for VGG networks. 
We expect that the difference between VGG and ResNet architectures (e.g., different dimensionality) will result in different EDNLs. 
Yet, the proximity between the observed EDNLs of these models validates that EDNL is indeed attributed to the task itself rather than the exact model. 
Interestingly, we could reduce the depth of such architecture despite the lack of residual connections. 
We accredit such success to the fact that our method can preserve the rich intermediate representations of the deeper original network during fine-tuning. 

\begin{figure}
	\centering
	\begin{minipage}{0.49\linewidth}
	    \includegraphics[width=0.99\textwidth, left]{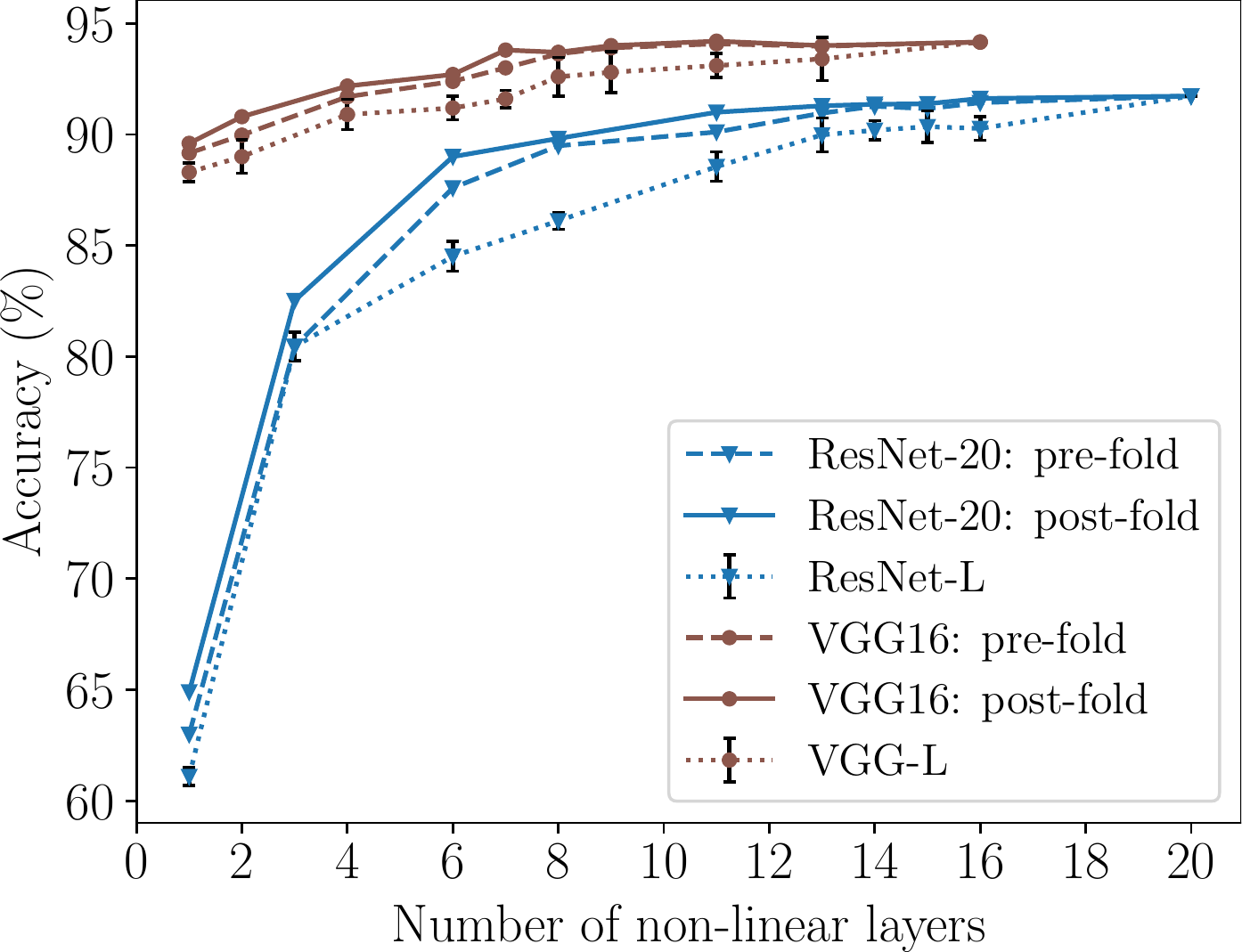}
		\label{5a_CIFAR10_ResNet_VGG_Pre_Post_Scratch}
	\end{minipage}
	\;
	\begin{minipage}{0.49\linewidth}
	    \includegraphics[width=0.99\textwidth, right]{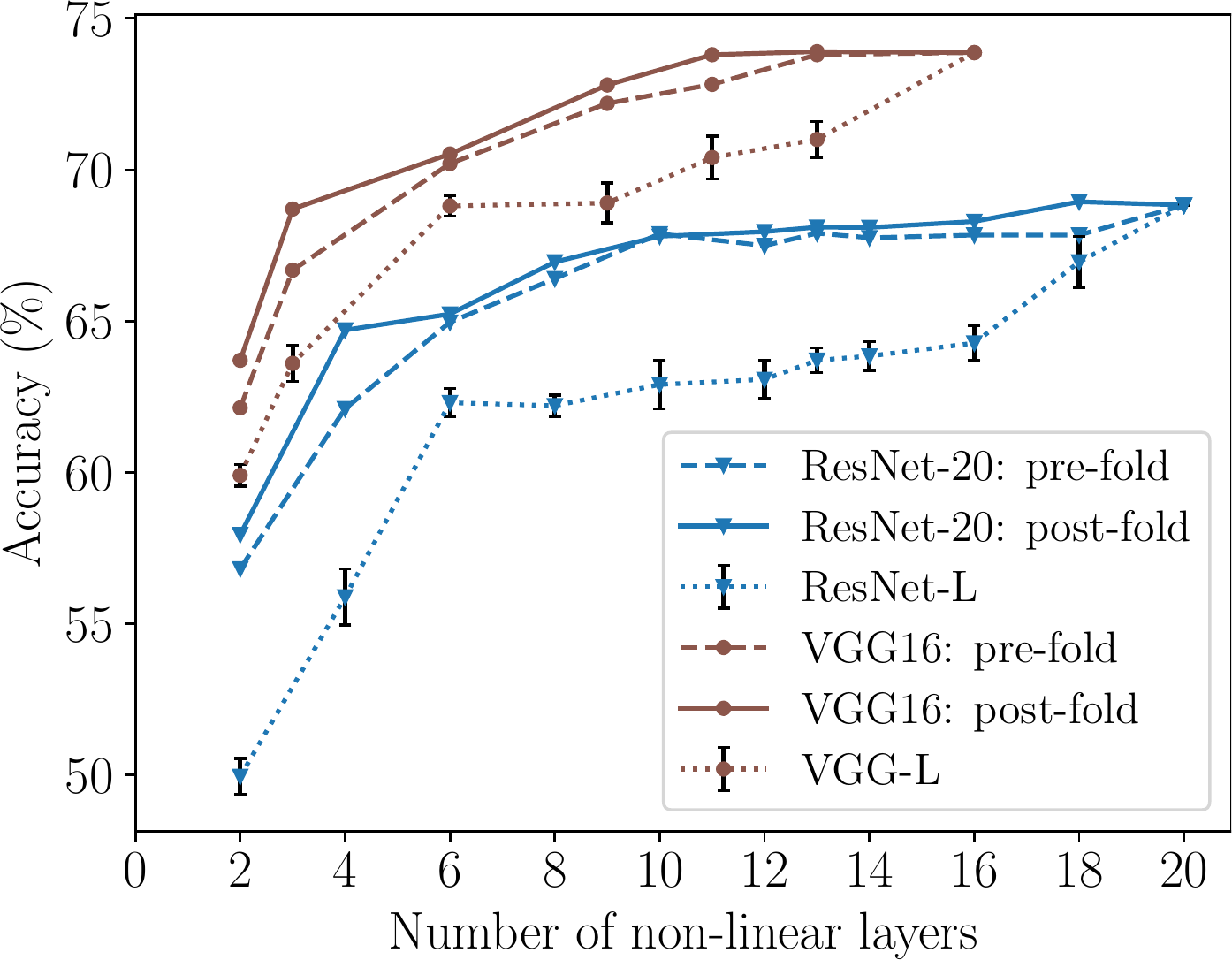}
		\label{5b_CIFAR100_ResNet_VGG_Pre_Post_Scratch}
    \end{minipage}	
    \caption{Relative contribution of pre-folding and post-folding. We perform pre-folding fine-tuning on ResNet-20 and VGG-16 and compare them to their shallower architectures resulting from post-folding. We also compare them to training the folded architectures from scratch, denoted by ResNet-L and VGG-L. Results are shown for CIFAR-10 (left) and CIFAR-100 (right).}
	\label{5_CIFAR_ResNet_VGG_Pre_Post_Scratch}
\end{figure}

In Figure \ref{5_CIFAR_ResNet_VGG_Pre_Post_Scratch} we compare networks obtained by pre-folding and post-folding. 
The results show that the EDNL of a folded network conforms to the one of the original network. 
We note that for folded networks, shallower architectures utilize more parameters, as the folding of two consecutive convolution layers with weights $\mathbf{W} \in \mathbb{R}^{3 \times 3 \times c \times c}$ result in a layer whose weights are $\mathbf{W} \in \mathbb{R}^{5 \times 5 \times c \times c}$, i.e., its size grows by $40\%$. 
Hence, and in contrast to MNIST, this experiment emphasizes the importance of depth even when it is disproportional to model's size, as networks were outperformed by deeper counterparts with fewer parameters. 
In addition, the slight accuracy increase of the folded networks quantifies the benefit of the post-folding phase. 
We believe that this makes them favorable to further efficiency optimization such as kernel size reduction. 
For example, we speculate that a $9 \times 9$ kernel resulted from folding four $3 \times 3$ kernels can be successfully distilled into a $7 \times 7$ kernel that nonetheless holds more capacity ($49$) than the four kernels altogether ($36$). 
We leave such optimization directions to future work.

We further compare our method to training randomly initialized networks with different depths as commonly practiced in NAS methods. 
Figure \ref{5_CIFAR_ResNet_VGG_Pre_Post_Scratch} shows the accuracy degradation of the folded architectures when trained from scratch rather than derived from their deeper source by our method. 

\begin{figure}
	\centering
	\begin{minipage}{0.49\linewidth}
	    \includegraphics[width=0.99\textwidth]{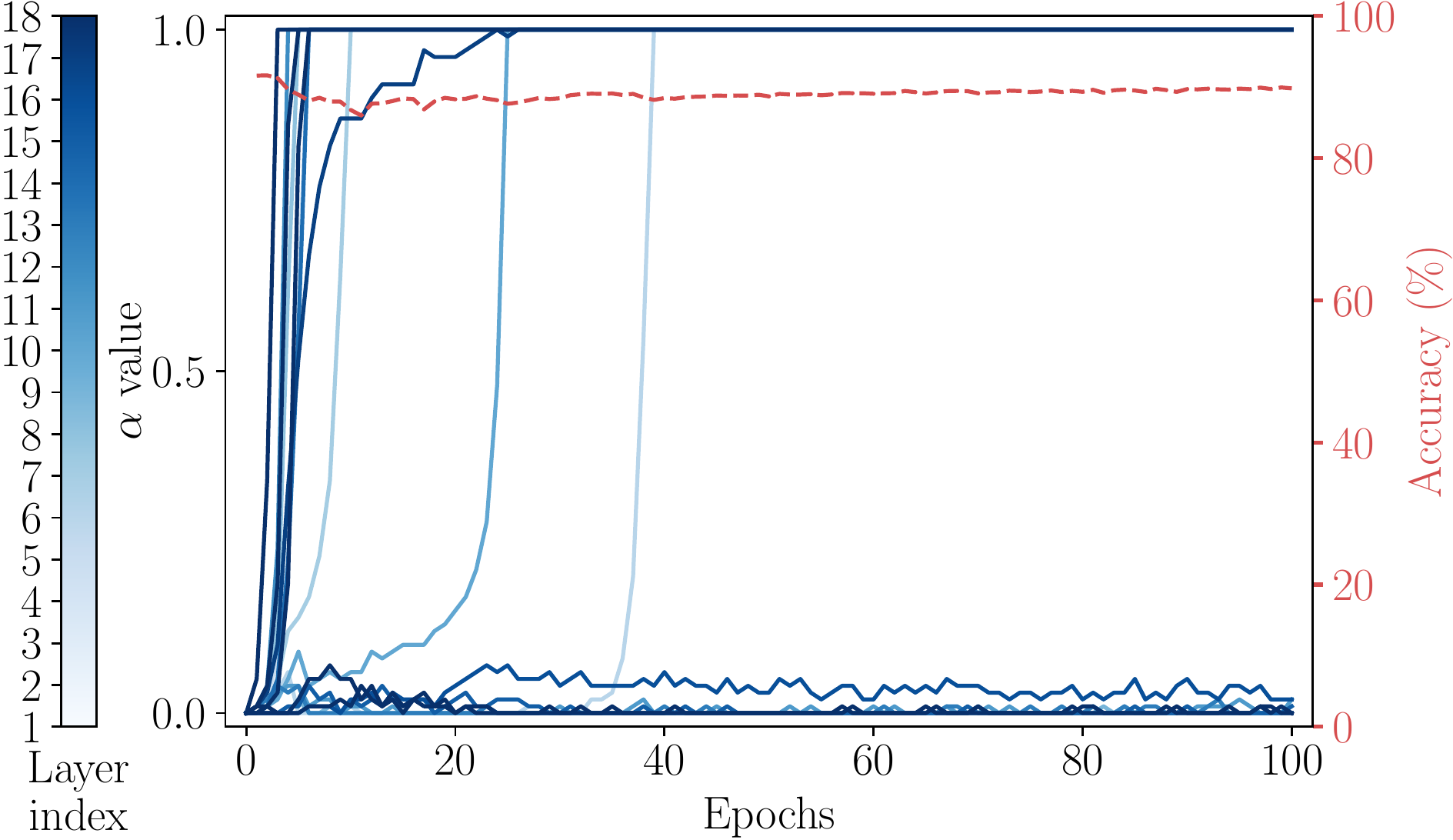}
	\end{minipage}
	\;
	\begin{minipage}{0.49\linewidth}
	    \includegraphics[width=0.99\textwidth]{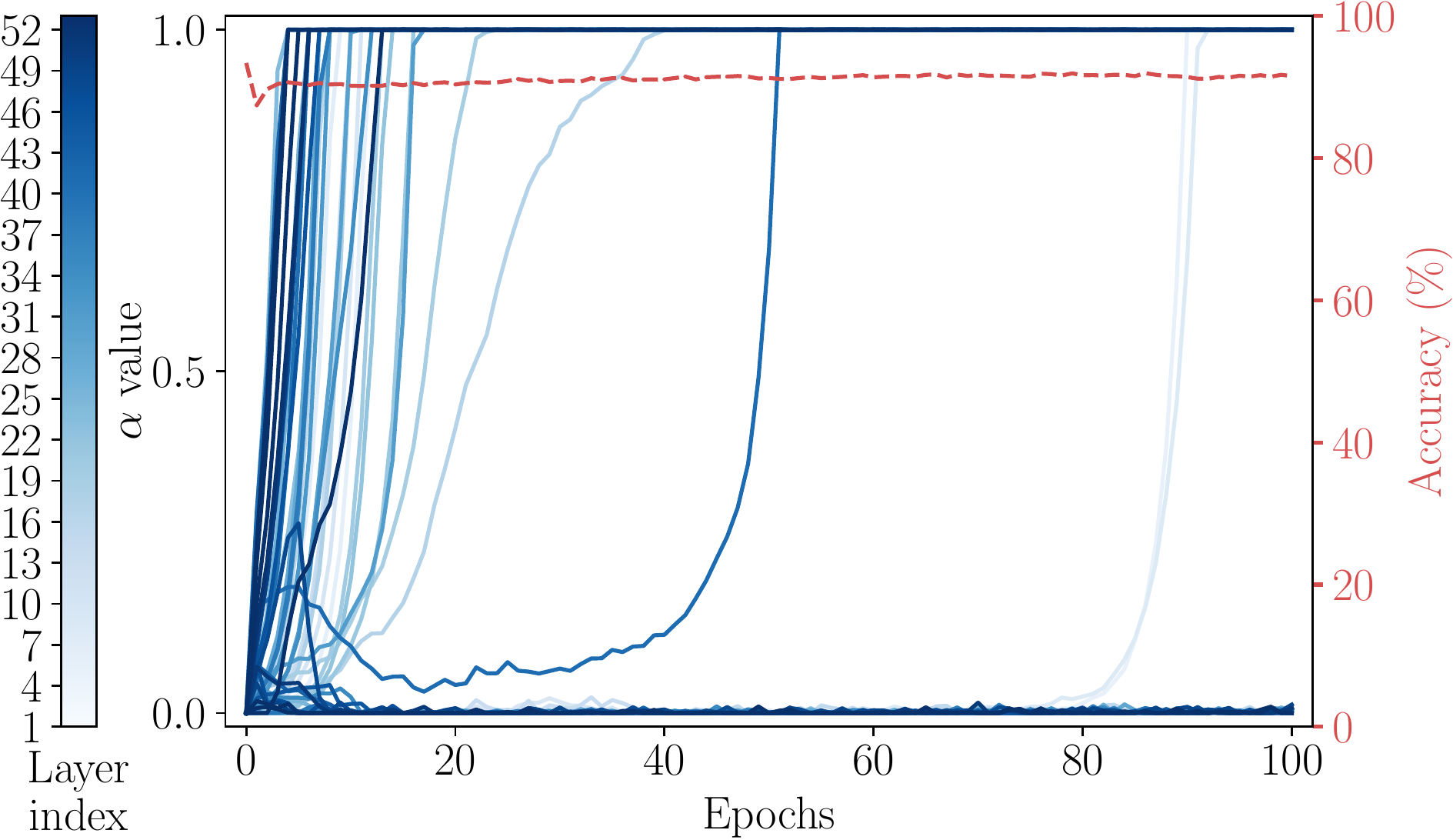}
	\end{minipage}
	\caption{Progression of $\alpha_l$ values corresponding to non-linear layers in ResNet-20 (left) and ResNet-56 (right) throughout the pre-folding phase with $\lambda_c=0.25$. As expected, all $\alpha$ values are either kept around zero or pushed to one.}	   	
	\label{8_alpha_prog}
\end{figure}

While Layer Folding can be applied with various loss functions in accordance with Equation \eqref{eq:generic_alpha_loss}, we show that our chosen loss function in Equation \eqref{eq:specific_alpha_loss} can effectively remove non-linear layers. 
Figure \ref{8_alpha_prog} validates that $\alpha$ values are indeed kept around zero or pushed to one, avoiding values in between. 
Additionally, we show that our method is not biased towards folding layers either in the beginning or the end of a given architecture. 
Table \ref{9_alpha_ind} shows the indices of the removed layers for folded ResNet and VGG networks such that their resulting depth corresponds to their EDNL.

\begin{table}
	\caption{Folded networks.}
	\vspace{1mm}
	\label{9_alpha_ind}
	\centering
	\resizebox{0.94\textwidth}{!}{
	\begin{tabular}{ c c c c c }
		\toprule
		Dataset & Model & Removed (white) and remaining (gray) activations  & Depth & Acc. (\%) \\
		\midrule
		\multirow{6}{*}{CIFAR-10} 
				         & ResNet-20 & \includegraphics[width=0.5\textwidth,height=3mm]{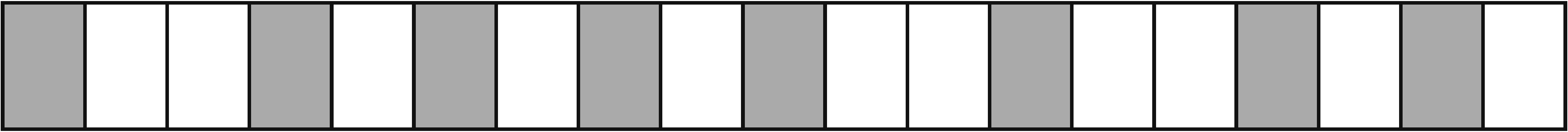} & 9 & 89.82 \\
				         & ResNet-32 & \includegraphics[width=0.5\textwidth,height=3mm]{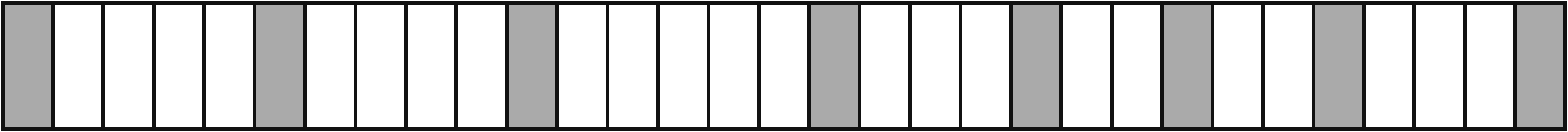} & 9 & 90.02 \\
				         & ResNet-44 & \includegraphics[width=0.5\textwidth,height=3mm]{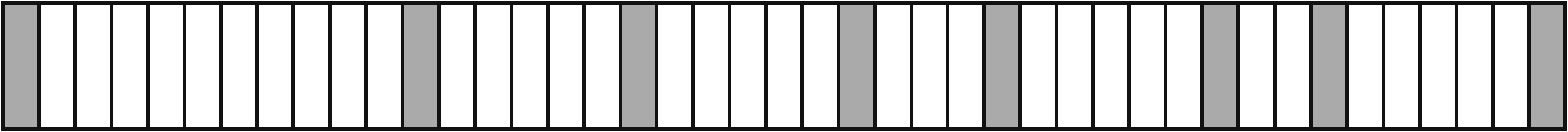} & 9 & 89.88 \\
				         & ResNet-56 & \includegraphics[width=0.5\textwidth,height=3mm]{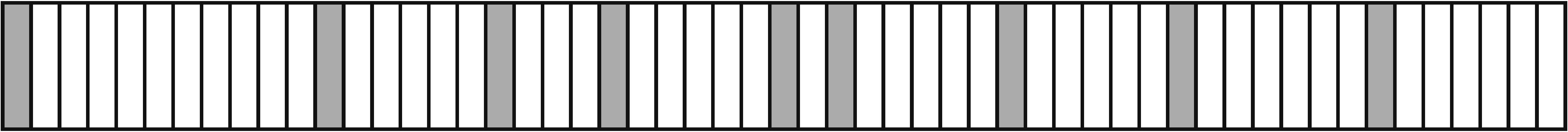} & 10 & 90.29 \\
		\cmidrule(l){2-5}
				         & VGG16     & \includegraphics[width=0.5\textwidth,height=3mm]{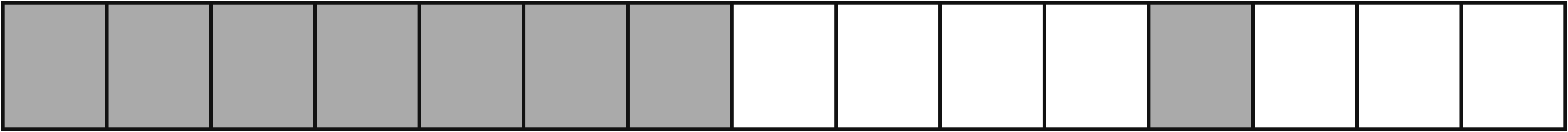} &  9 & 93.89 \\
				         & VGG19     & \includegraphics[width=0.5\textwidth,height=3mm]{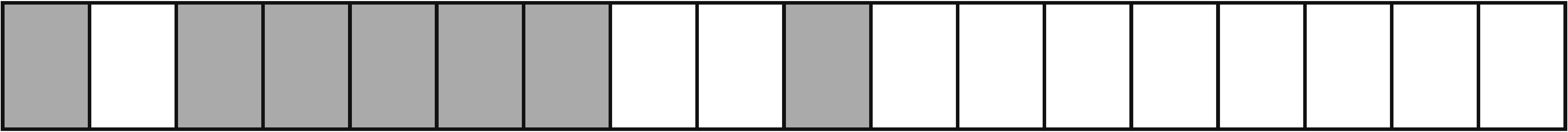} &  8 & 93.23 \\
		\midrule
		\multirow{6}{*}{CIFAR-100} 				
				          & ResNet-20 & \includegraphics[width=0.5\textwidth,height=3mm]{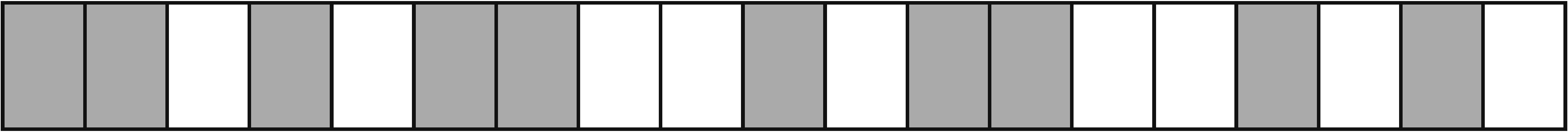} & 11 & 67.88 \\
				          & ResNet-32 & \includegraphics[width=0.5\textwidth,height=3mm]{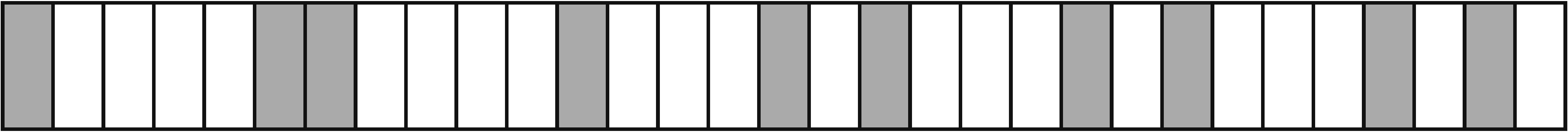} & 11 & 68.20 \\
				          & ResNet-44 & \includegraphics[width=0.5\textwidth,height=3mm]{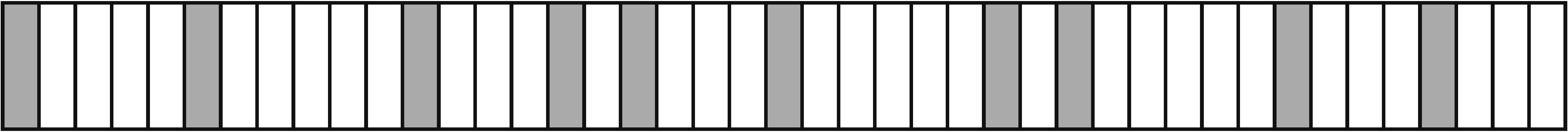} & 11 & 67.96 \\
				          & ResNet-56 & \includegraphics[width=0.5\textwidth,height=3mm]{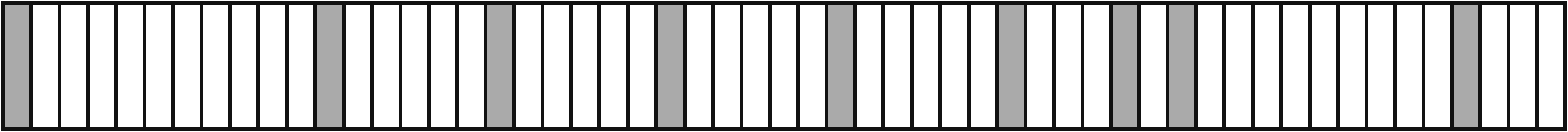} & 10 & 67.04 \\
		\cmidrule(l){2-5}
				          & VGG16     & \includegraphics[width=0.5\textwidth,height=3mm]{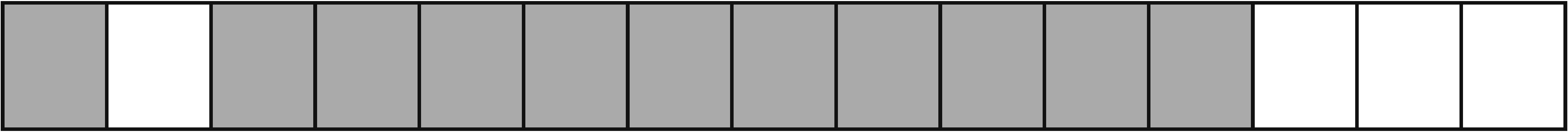} & 12 & 72.82 \\
				          & VGG19     & \includegraphics[width=0.5\textwidth,height=3mm]{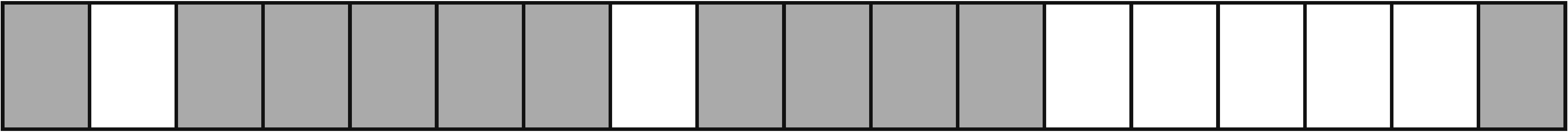} & 12 & 73.18 \\
		\bottomrule
	\end{tabular} }
\end{table}

\section{Depth Optimization for Efficient Networks}
\label{optimization_section}
In this section we utilize our method to optimize networks with respect to both accuracy and efficiency. 
We perform our experiments on the ImageNet image classification task \citep{ILSVRC15} and measure the latency of all models on NVIDIA Titan X Pascal GPU. 


We consider the commonly used MobileNetV2 \citep{sandler2018mobilenetv2} and EfficientNet-lite \citep{liu2020higher}.
We focus on these models for their attractiveness for hardware and edge devices, mostly credited to their competitive latency and the exclusion of squeeze-and-excite layers \citep{hu2018squeeze} employed by other state-of-the-art networks. 

Both MobileNetV2 and EfficientNet-lite consist of multiple mobile inverted bottleneck blocks (MBConv). 
An MBConv block is composed of three convolution layers and two activations: 
(1) an expansion layer with $\mathbf{W} \in \mathbb{R}^{1 \times 1 \times c \, t \times c}$ where $t$ denotes an expansion factor over the input channel dimension $c$ followed by ReLU6, 
(2) a depthwise layer with $\mathbf{W} \in \mathbb{R}^{3 \times 3 \times c}$ followed by ReLU6, and 
(3) a projection layer with $\mathbf{W} \in \mathbb{R}^{1 \times 1 \times c \times c \, t}$. 
Removing only the first ReLU6 will allow us to fold the expansion and depthwise layers into a convolution with $\mathbf{W} \in \mathbb{R}^{3 \times 3 \times c \, t \times c}$. 
Interestingly, MobileNetEdgeTPU \citep{howard2019searching} followed this exact approach and adopted fused inverted bottleneck. 
In this model, expansion layers were folded with depthwise layers despite a FLOPs increase, acknowledging the potential latency reduction. 
This was performed on the the first MBConvs. 
Indeed, in the general case, such folding may lead to a disadvantageous increase in FLOPs when $c \gg 9$. 
Removing the second ReLU6 will result in a similar FLOPs increase. 
However, we recognize that removing both ReLU6 activations will allow folding all three convolutions into a single layer with $\mathbf{W} \in \mathbb{R}^{3 \times 3 \times c \times c}$. 
This effectively halve the computational load for blocks in MobileNetV2 where $t=6$. 
In our experiments, we leverage this favorable scenario by forcing folding of an entire MBConv blocks. 
For every block, we share the same $\alpha$ among both of its activations. 
This ensures that they are either removed together or remain. 

\begin{figure}
	\centering
	\begin{minipage}{0.47\linewidth}
	    \includegraphics[width=0.99\textwidth]{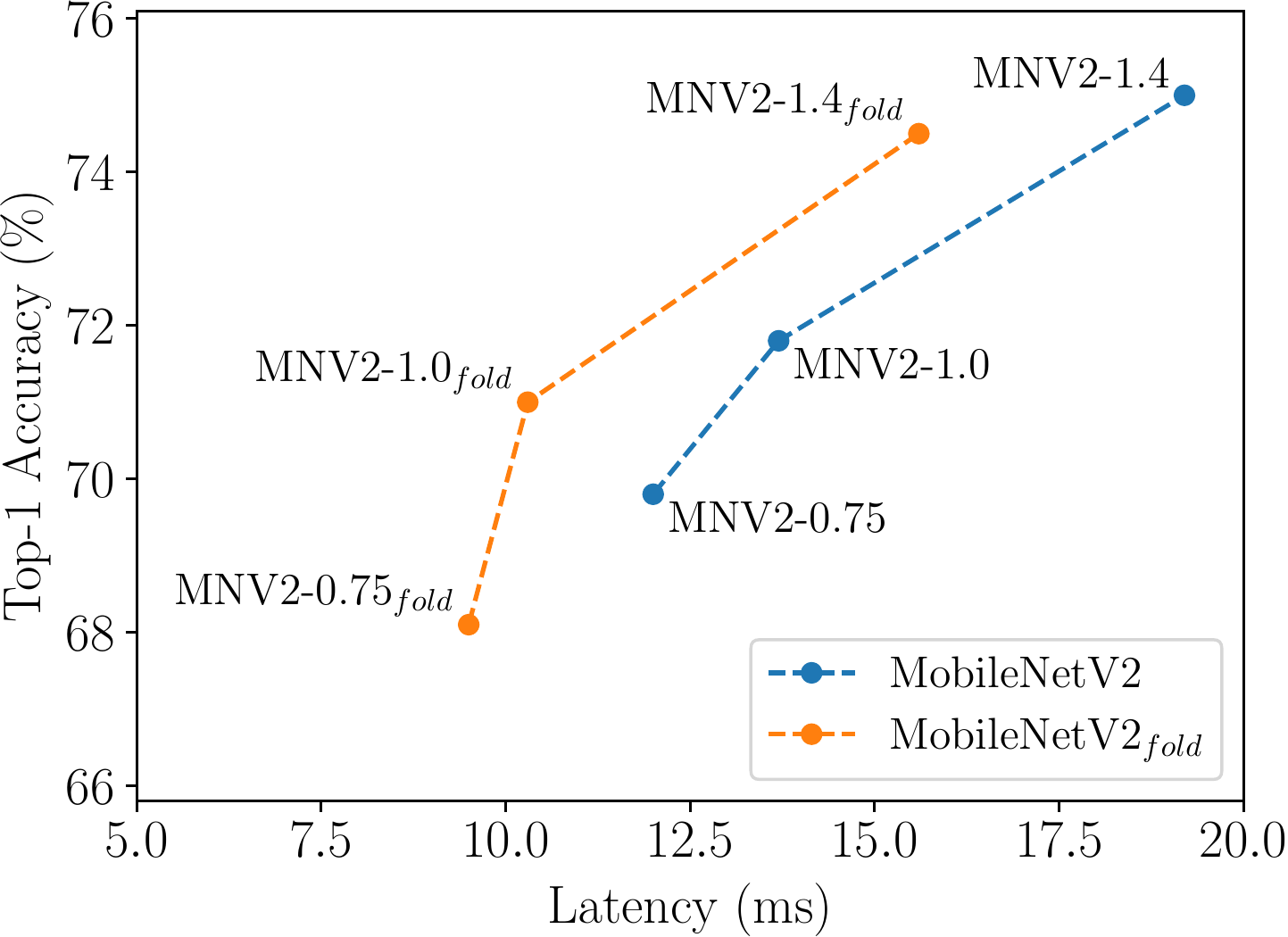}
	    \caption{Layer Folding results on ImageNet.}
   		\label{6_ImageNet_MNV2}
	\end{minipage}
	\;
	\begin{minipage}{0.51\linewidth}
		\captionof{table}{Latency and FLOPs reduction obtained by applying Layer Folding on MobileNetV2 (MNV2) and EfficientNet (EffNet) on ImageNet.}
		\label{ImageNet_table}
		\centering
		\renewcommand{\arraystretch}{1.25}
		\resizebox{\textwidth}{!}{%
		\begin{tabular}{ c c c c }
			\toprule
			\multirow{2}{*}{Model} & Acc. (\%) /    & Latency   & FLOPs \\
			                       & Acc. Drop (\%) & Reduction & Reduction \\
			\midrule
			MNV2-0.75 		& 68.1 / 1.7 & 21\% & 4\% \\
			MNV2-1.0 		& 71.0 / 0.8 & 25\% & 7\% \\
			MNV2-1.4 		& 75.5 / 0.5 & 19\% & 3\% \\
			EffNet-lite0 	& 74.6 / 0.5 & 15\% & 3\% \\
			EffNet-lite1 	& 75.8 / 1.0 & 13\% & 0\% \\
			\bottomrule
		\end{tabular} %
		}
    \end{minipage}	
\end{figure}

We apply Layer Folding on MobileNetV2-1.4, MobileNetV2-1.0, MobileNetV2-0.75, EfficientNet-lite0 and EfficientNet-lite1. 
Our implementation details are provided in Appendix \ref{implementation_details}. 
As shown in Figure \ref{6_ImageNet_MNV2}, our folded models outperform MobileNetV2 variants. 
For example, compared to MobileNetV2-0.75, we obtain a model with a $1.2\%$ higher top-1 accuracy at a $14\%$ faster execution time. 
Table \ref{ImageNet_table} shows the latency reduction obtained for MobileNet and EfficientNet models. 
These improvements should be viewed while taking the simplicity and efficiency of our method into account. 

In the experiment above, we improve networks' latency mainly by revisiting their activations. 
We suggest that other optimization techniques that operate on different architectural components, such as pruning, can be used conjointly with our method. 
That is, that their expected contribution is additive rather than alternative to ours as they focus on width and remove weights while our method focuses on depth and removes activations. 

\section{Conclusion}
\label{conclusion_section}
In this work we propose a novel method for removing non-linear activations. 
Extensive experiments on several image classification tasks show that our method can significantly reduce the depth of neural networks with a minor effect on accuracy. 
We find that there is a minimal number of non-linear layers to which networks can be reduced while retaining accuracy which we denote as EDNL. 
We show that networks that vary by depth and architecture share a similar EDNL, suggesting that EDNL is an attribute of a desired mapping over a certain task rather than a specific model. 
Our work provides empirical results that verify and bridge previous theoretical works on the importance of depth and recent works that reason the profitability of increased depth. 
The scope of this work is EDNL evaluation of CNNs with ReLU activations. 
We leave further study of EDNL and the natural extension to other architectures for future work.
Finally, we show how reduced depth can aid latency reduction on hardware devices and provide efficient alternatives to mobile network architectures. 

\section{Broader Impact}
\label{impact_section}
The main positive impact of our work is the potential reduction of energy consumption which aids environmental protection. 
This is reflected in three aspects; first, our method allows to reduce the latency and power consumption of neural network-based applications. 
Second, compared to NAS methods that require training multiple models, our method requires fine-tuning of a single model, reducing the cost of the optimization process. 
Third, our findings on EDNL might sprout additional means to quantify and bound the expressiveness of neural networks. 
Additionally, our method allows the deployment of neural networks on computationally-constrained edge-devices that would not be possible otherwise, along with its positive and negative societal consequences. 

\bibliography{layer_folding_v1}

\appendix

\section{Additional Results}
\label{ablation}

\begin{figure}[h]
\centering
    \begin{subfigure}{0.48\linewidth}
		\includegraphics[width=\linewidth]{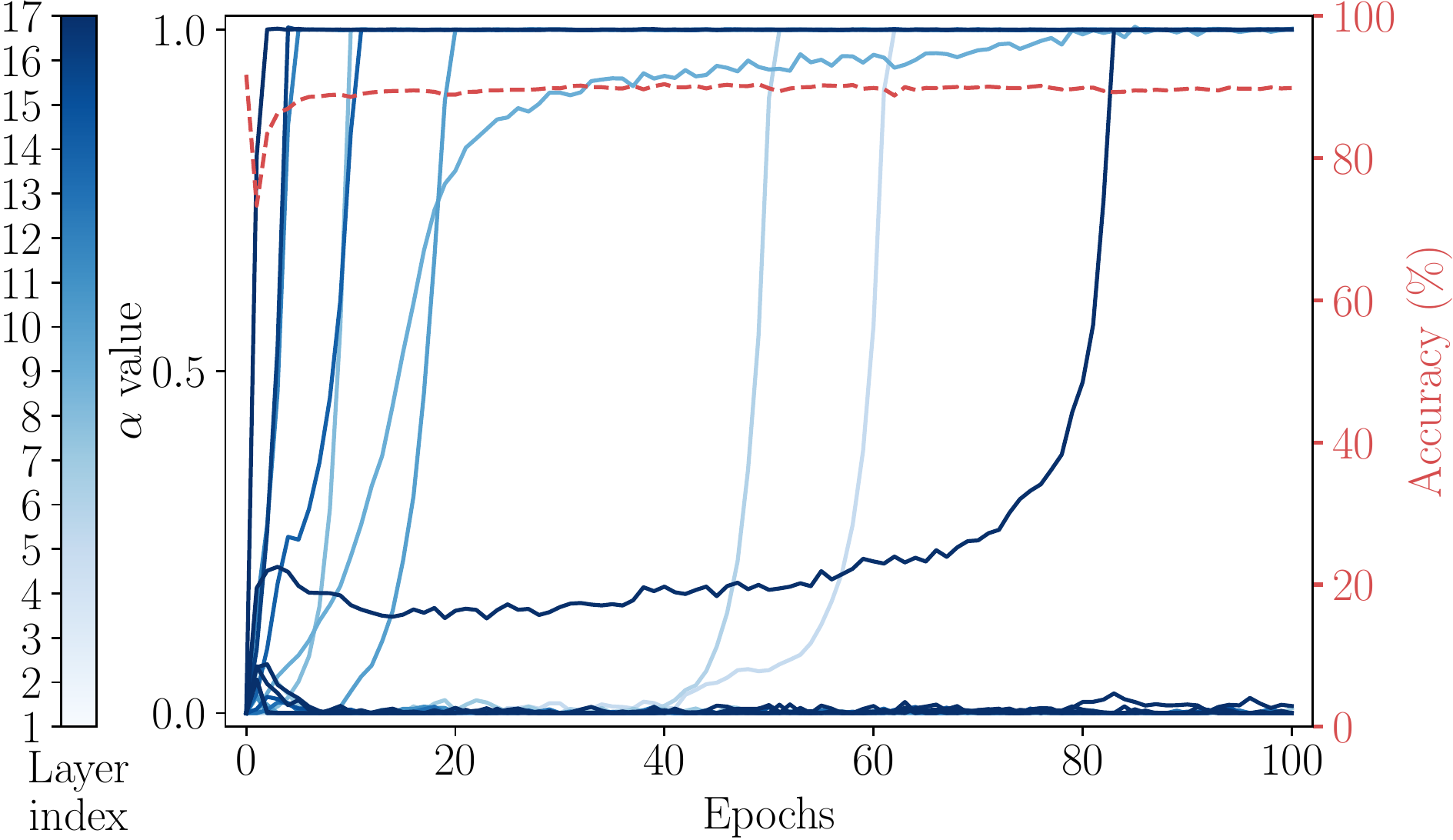} 
    		\caption{ResNet-20, $\lambda_c = 0.1$}
    \end{subfigure}\hfill
    \begin{subfigure}{0.48\linewidth}
		\includegraphics[width=\linewidth]{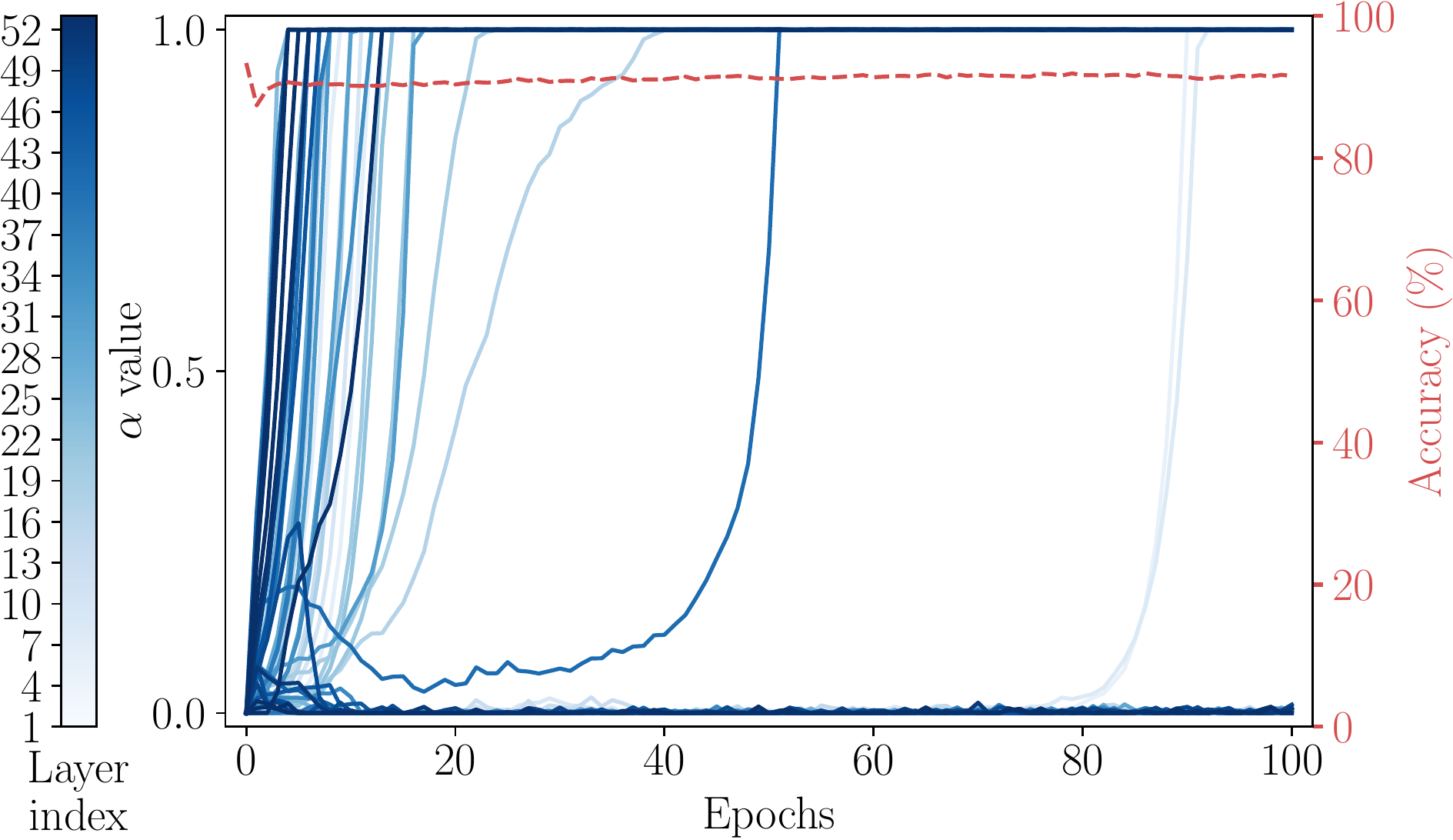}
    		\caption{ResNet-56, $\lambda_c = 0.1$}
	\end{subfigure}

    \begin{subfigure}{0.48\linewidth}
		\includegraphics[width=\linewidth]{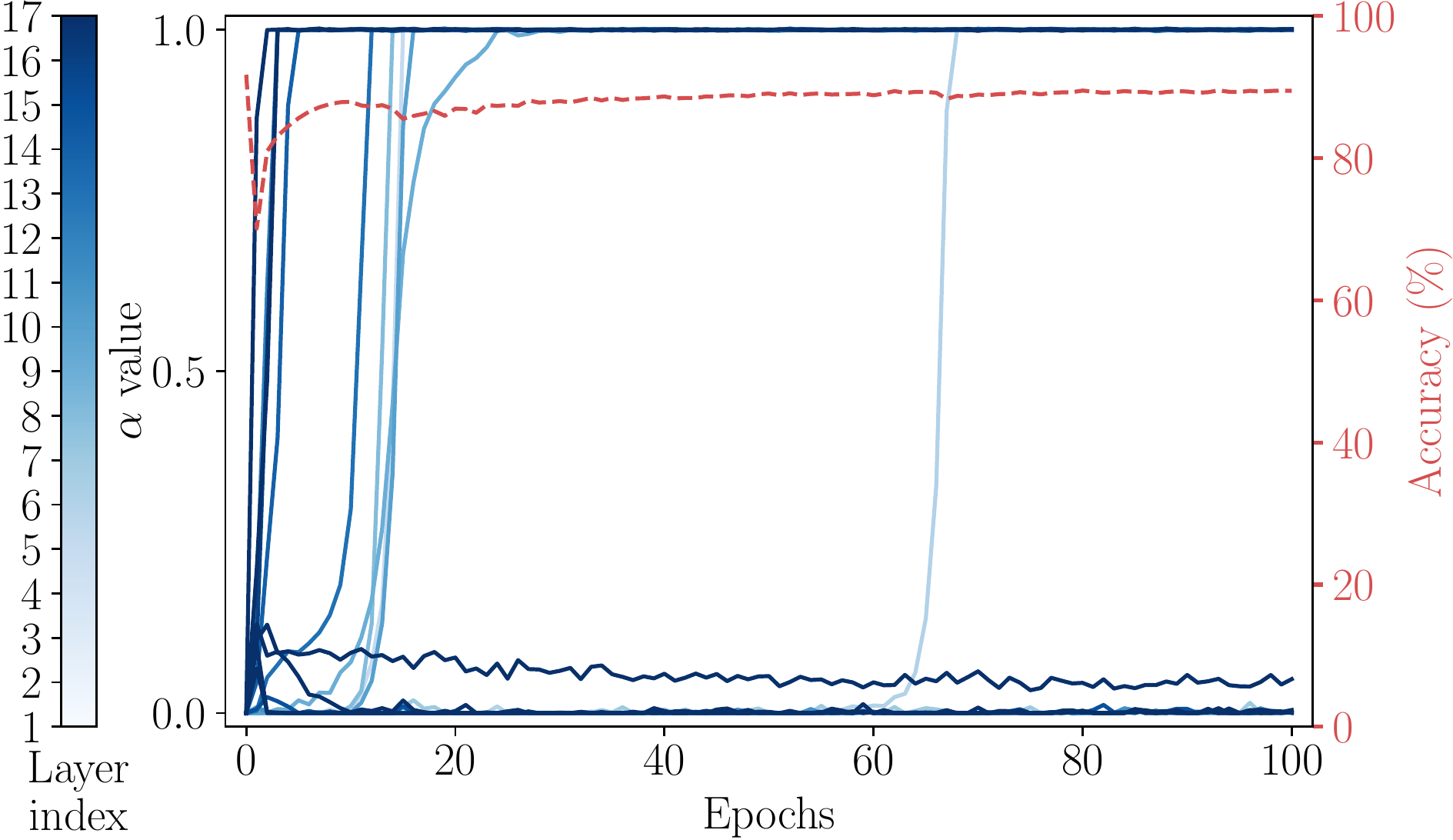} 
    		\caption{ResNet-20, $\lambda_c = 0.2$}
    \end{subfigure}\hfill
    \begin{subfigure}{0.48\linewidth}
		\includegraphics[width=\linewidth]{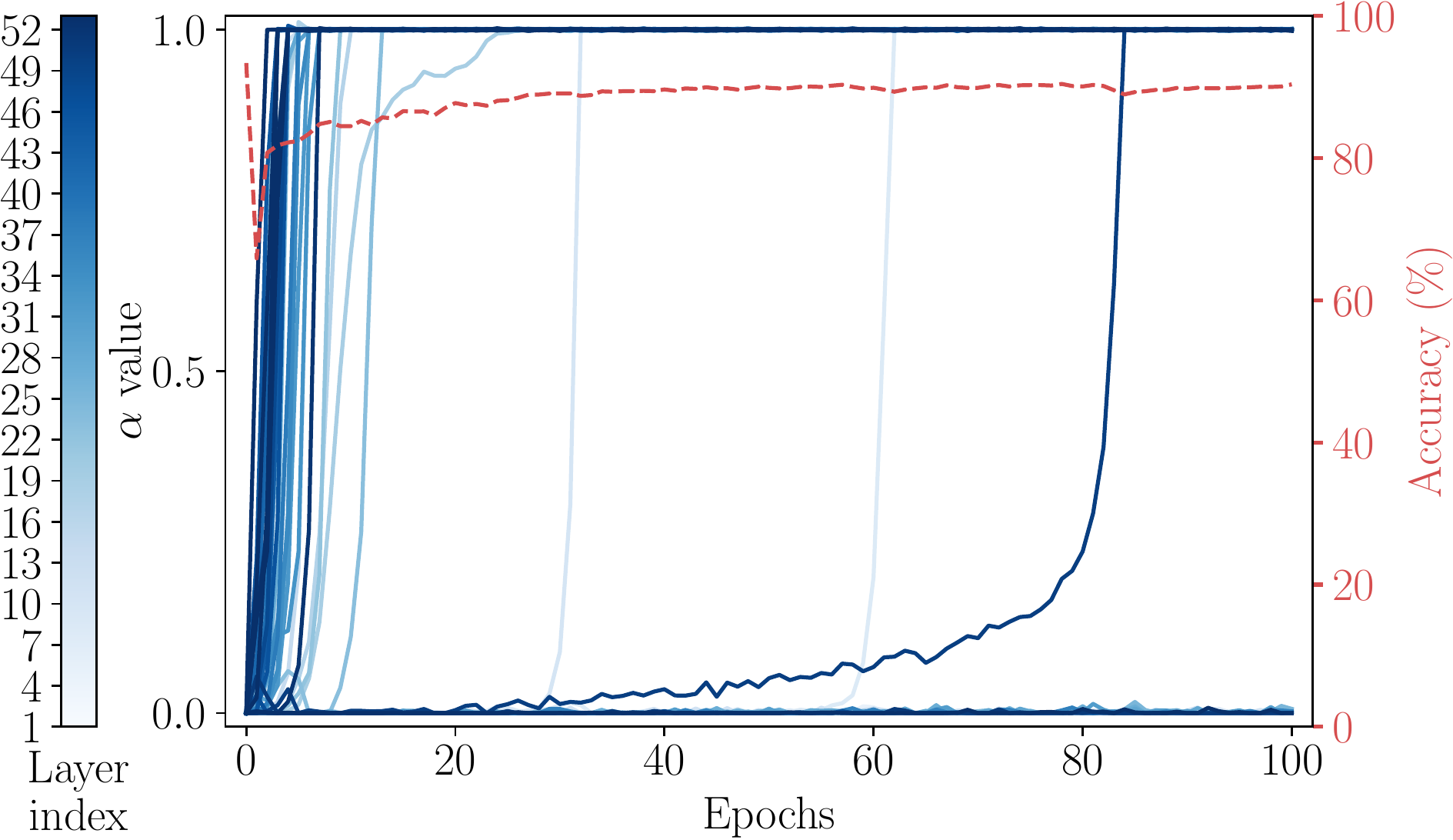}
    		\caption{ResNet-56, $\lambda_c = 0.2$}
	\end{subfigure}
	
	    \begin{subfigure}{0.48\linewidth}
		\includegraphics[width=\linewidth]{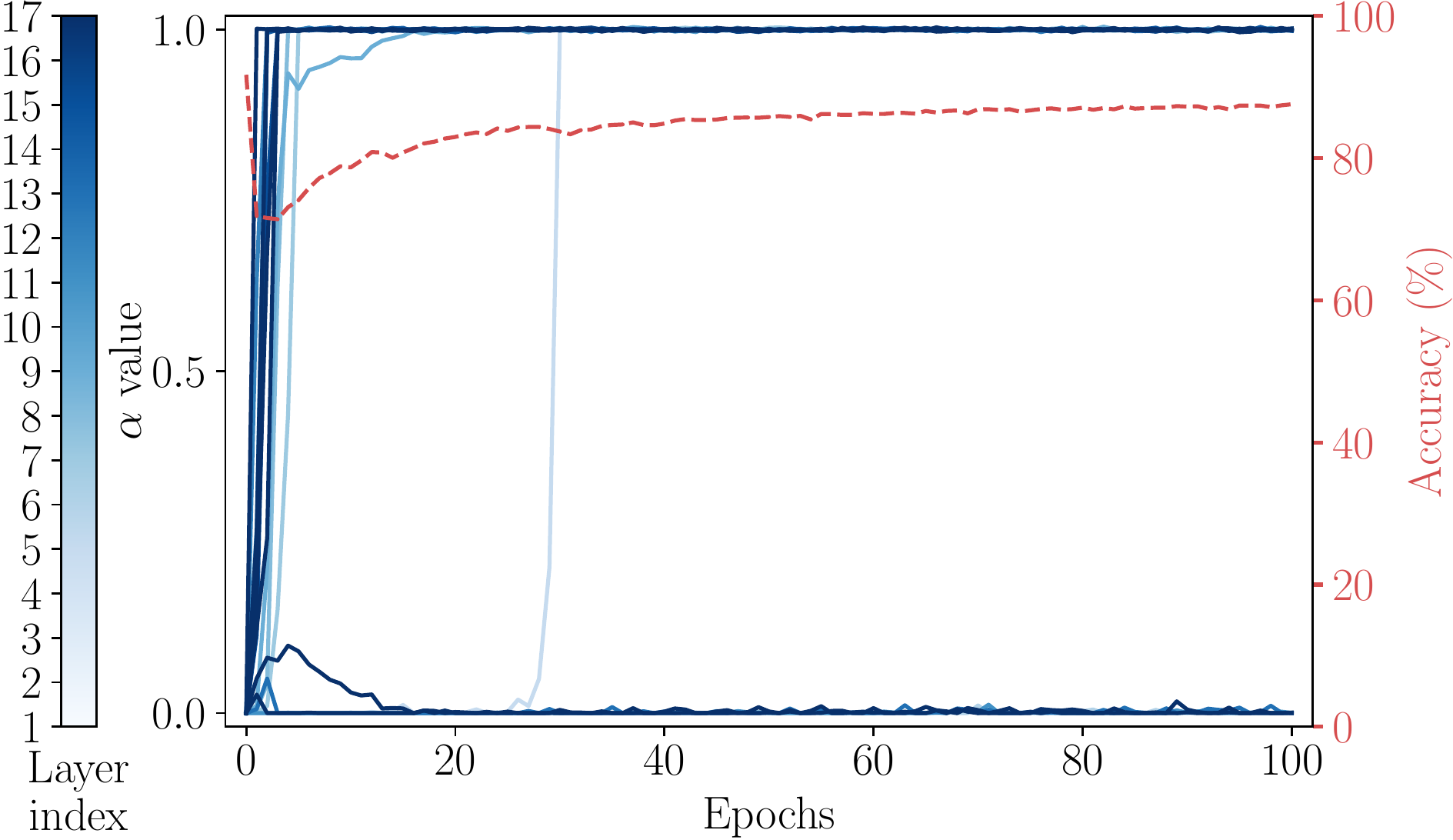} 
    		\caption{ResNet-20, $\lambda_c = 0.4$}
    \end{subfigure}\hfill
    \begin{subfigure}{0.48\linewidth}
		\includegraphics[width=\linewidth]{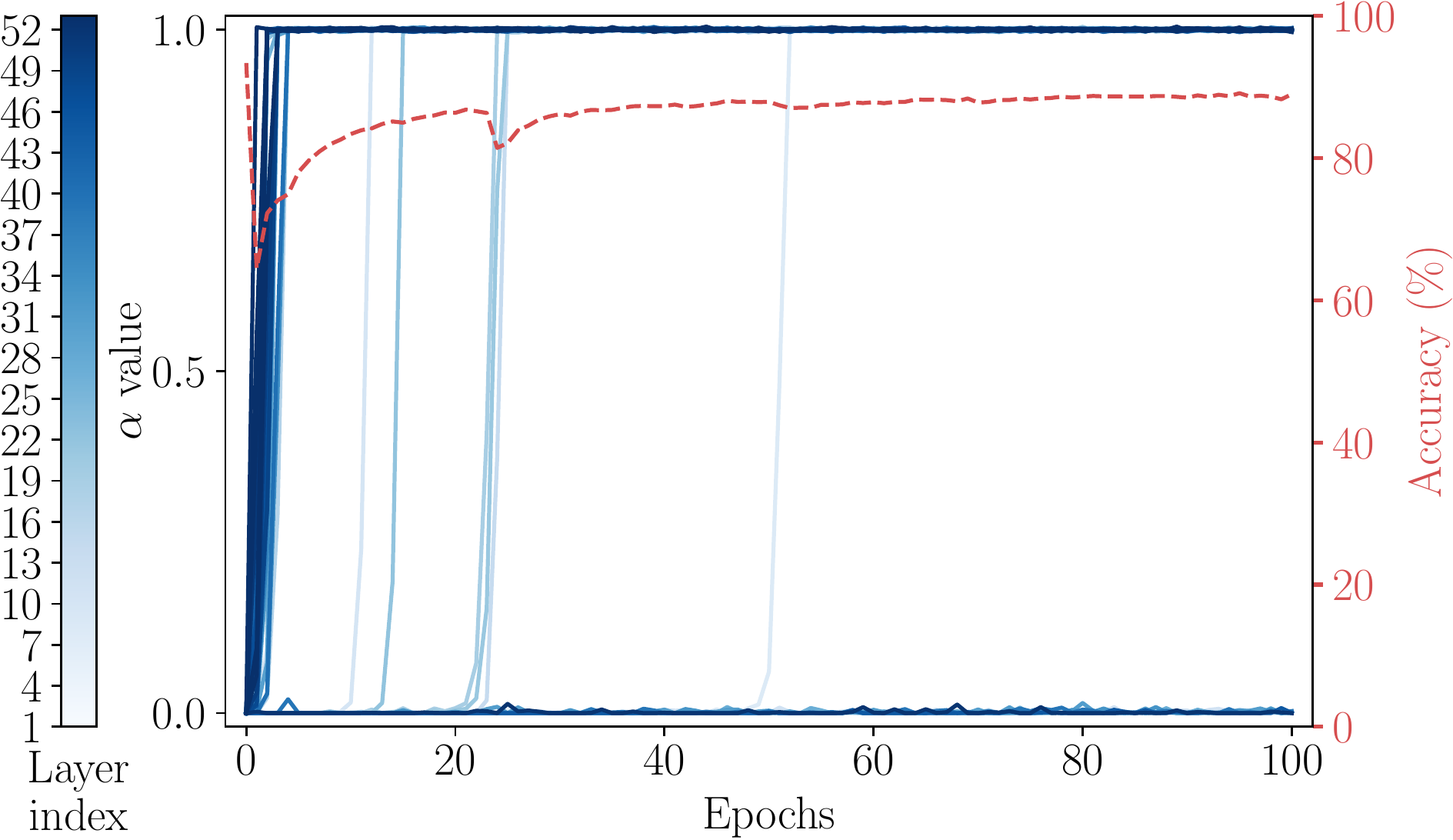}
    		\caption{ResNet-56, $\lambda_c = 0.4$}
	\end{subfigure}
	\caption{Layer Folding with different values of $\lambda_c$.}
    \label{alpha_prog_normal}
\end{figure}

\begin{figure}
\centering
    \begin{subfigure}{0.48\linewidth}
		\includegraphics[width=\linewidth]{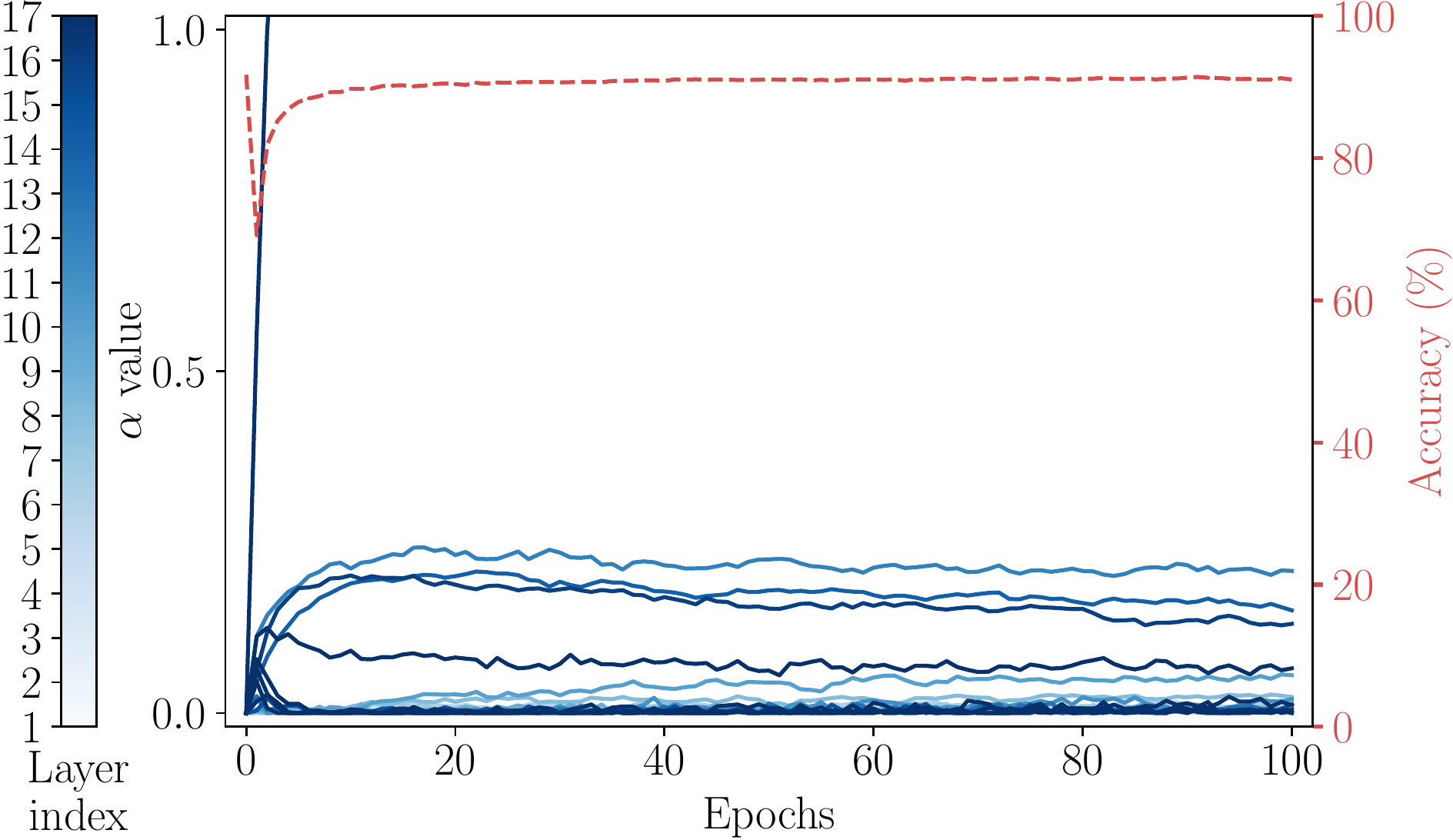} 
    		\caption{ResNet-20, $\lambda_c = 0.01$}
    \end{subfigure}\hfill
    \begin{subfigure}{0.48\linewidth}
		\includegraphics[width=\linewidth]{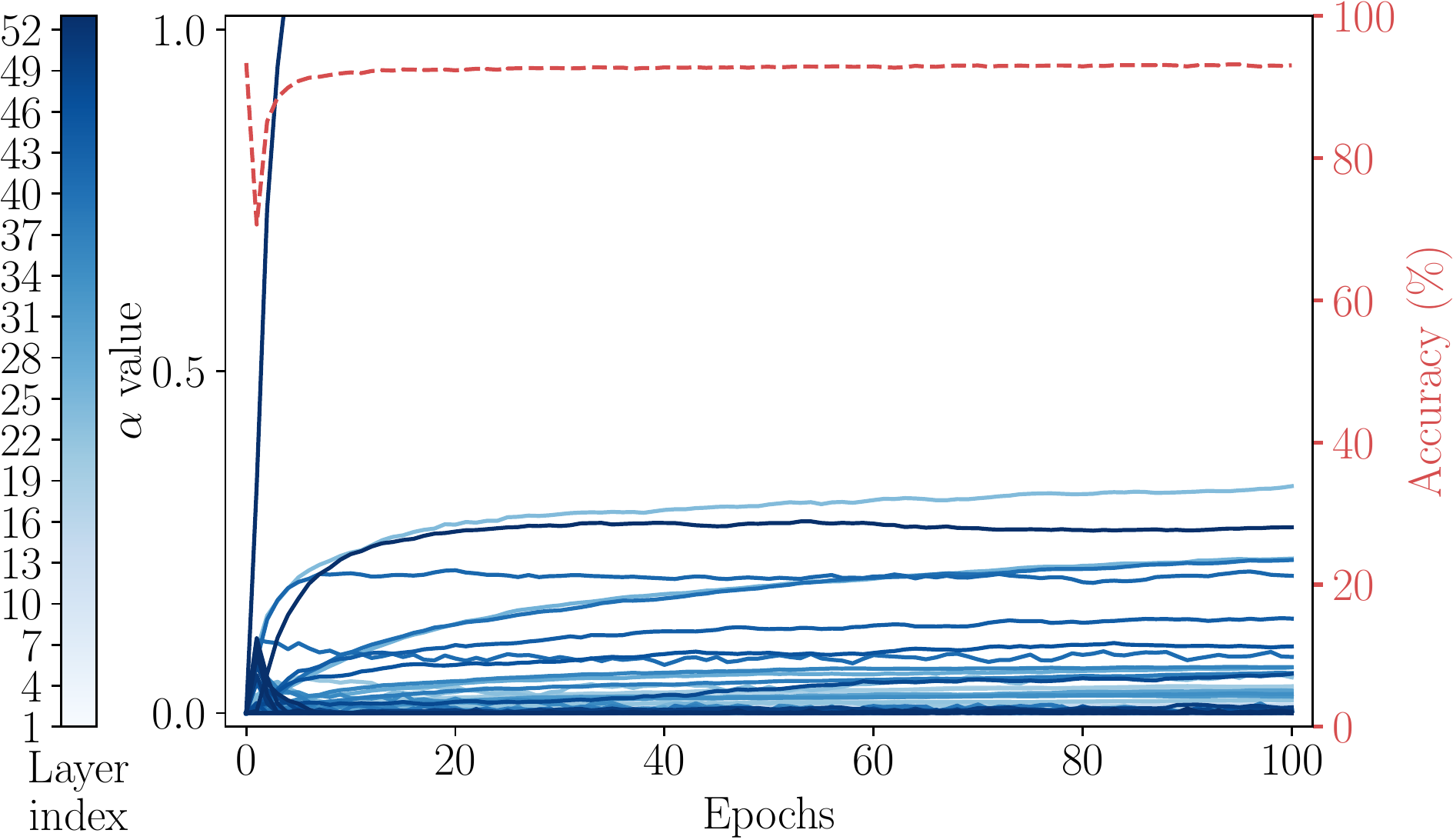}
    		\caption{ResNet-56, $\lambda_c = 0.01$}
	\end{subfigure}

    \begin{subfigure}{0.48\linewidth}
		\includegraphics[width=\linewidth]{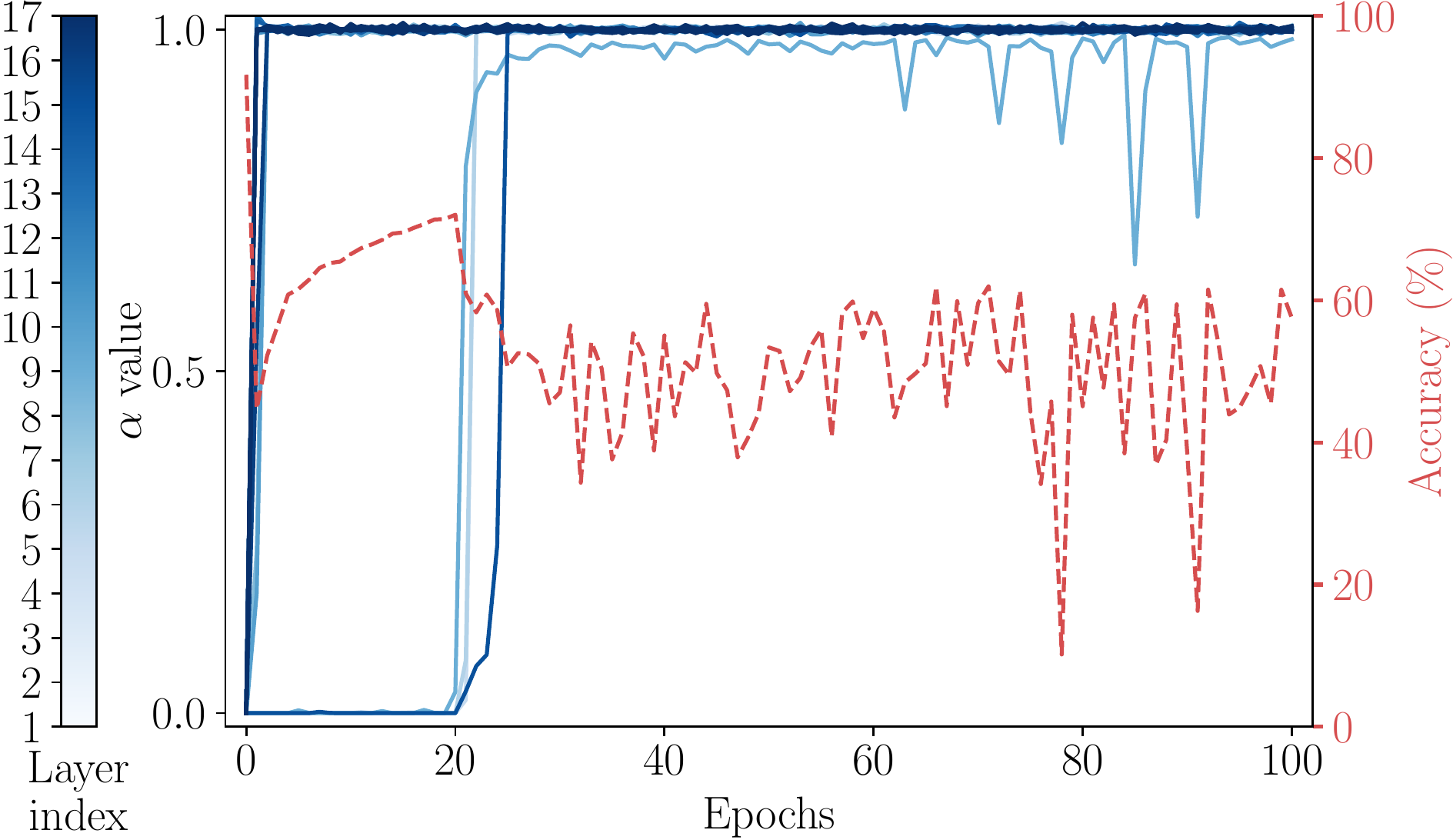} 
    		\caption{ResNet-20, $\lambda_c = 1$}
    \end{subfigure}\hfill
    \begin{subfigure}{0.48\linewidth}
		\includegraphics[width=\linewidth]{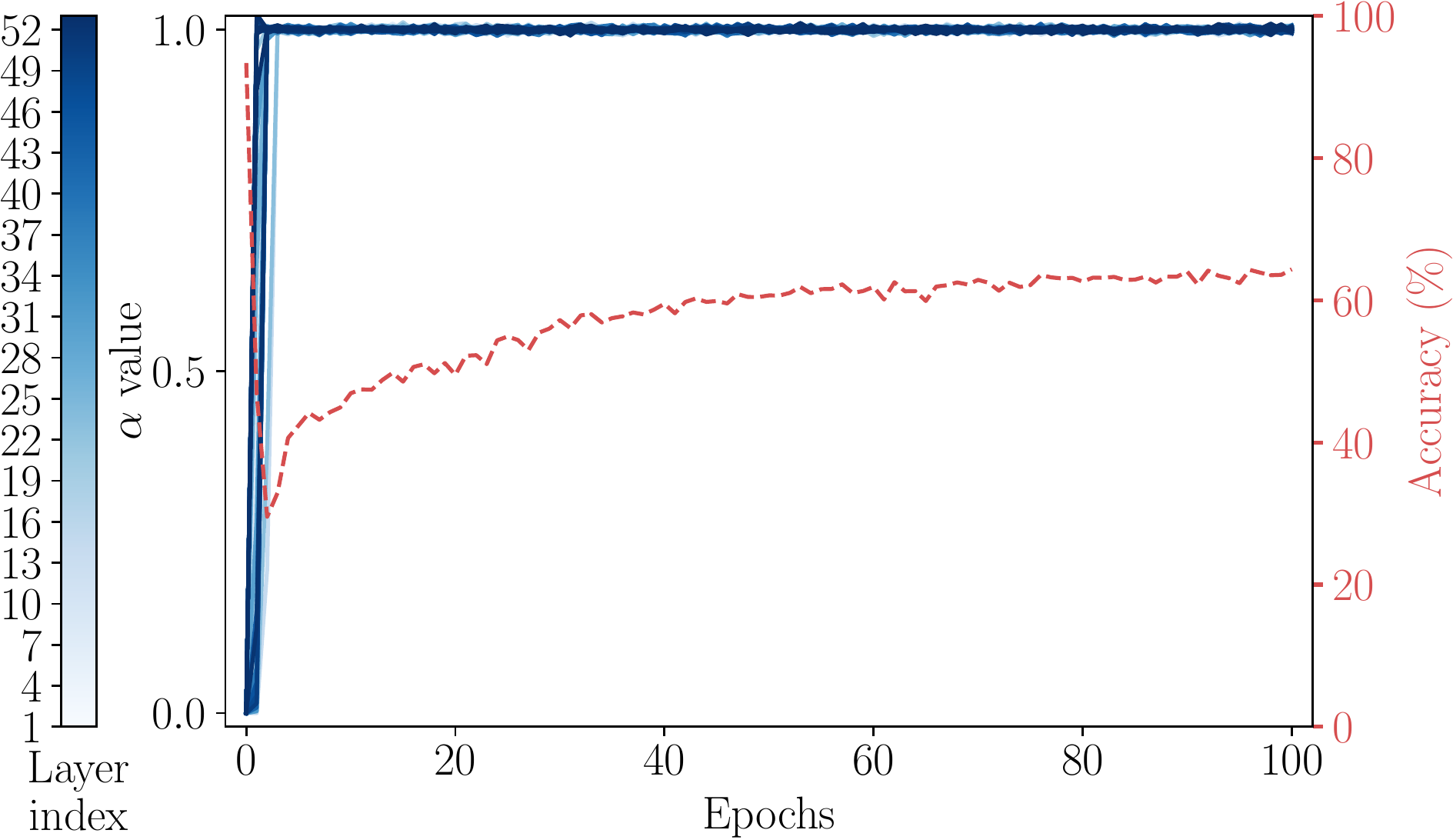}
    		\caption{ResNet-56, $\lambda_c = 1$}
	\end{subfigure}
	\caption{Layer Folding with extreme values of $\lambda_c$ such that none (top) or all (bottom) of the non-linear layers are removed.}
    \label{alpha_prog_unnormal}
\end{figure}

Figure \ref{alpha_prog_normal} shows the progression of $\alpha_l$ values  on ResNet-20 and ResNet-56 during the course of the pre-folding phase for different values of $\lambda_c$. 
As can be seen, our method effectively remove non-linear layers while maintaining accuracy, validating our choice of $h$ (see Equation \eqref{eq:specific_alpha_loss}, Section \ref{method_section}). 
In addition, the resulting $\alpha$ values when pre-folding concludes are either 0 or 1, indicating that $\tau$ values do not need special tuning. 
In Figure \ref{alpha_prog_unnormal} we show extreme cases where $\lambda_c$ values are set such that none or all of the non-linear layers are removed. 

\section{Implementation Details}
\label{implementation_details}

For MNIST, we use fully-connected networks with depth $L \in [2:10]$, ReLU activation and width $d=256$ for all layers. 
We train these networks for 10 epochs with a learning rate of $0.1$ and SGD with momentum of 0.9. 
For CIFAR-10 and CIFAR-100, we use pre-trained models of ResNet-20, ResNet-32, ResNet-44, ResNet-56, VGG16 and VGG19 from \citep{chenyaofo2019}. 
We apply Layer Folding with $c_l=1$, $l=1:L$, $p=2$, $\tau=0.9$. 
We note that these values were chosen for simplicity, and that from our experiments larger values of $p$ and $\tau$ provide similar results. 
We use $\lambda_c = 1$ and multiply it by $2^{0.5}$ ($2^{-0.5}$) to increase (decrease) the number of non-linear layers being removed. 
The networks above were trained for 200 epochs, with a starting learning rate of 0.1 which is reduced by one-tenth at the 100th and 150th epochs. 
Hence, we run the pre-folding and post-folding fine-tuning phases for 100 epochs each with a learning rate of $10^{-3}$. 
When training from scratch folded ResNet and VGG models (see Figure \ref{5_CIFAR_ResNet_VGG_Pre_Post_Scratch}, Section \ref{EDNL_section}), we match the training schedule of the folded networks, i.e., a starting learning rate of 0.1 which is reduced by one-tenth at the 100th and 150th epochs of the total 400 epochs. 
As training depends on initialization, we repeat the training for five times and take the average test accuracy. 
We implemented our code using PyTorch \citep{DBLP:journals/corr/abs-1912-01703} and NVIDIA GeForce GTX 1080Ti. 
In these settings, both pre-folding and post-folding phases take less than an hour for each of the above networks.


For ImageNet, we use MobileNetV2-1.4, MobileNetV2-1.0, MobileNetV2-0.75, EfficientNet-lite0 and EfficientNet-lite1. 
For all models we use augmentations drawn from the models’ original training flow - standard augmentations for MobileNetV2 and AutoAugment \citep{DBLP:conf/cvpr/CubukZMVL19} for EfficientNet-lite. 
We use Adam optimizer ($\beta_1 = 0.9$, $\beta_2 = 0.999$), $0.1$ label smoothing and batch size of $64$. 
We use weight decay of $2\mathrm{e}{-5}$. 
During the pre-folding phase we use a learning rate of $\mathrm{1e-4}$ and during the post-folding phase we use a cosine learning rate decay from $1\mathrm{e}{-5}$ to $1\mathrm{e}{-6}$. 
For MobileNetV2 variants we use exponential moving average with a decay factor of $0.9999$. 
The pre-folding phase consists of $5$ epochs and the post-folding phase consists of $20$ epochs.
For EfficientNet-lite we use Stochastic Depth \citep{DBLP:conf/eccv/HuangSLSW16} with survival probability of $0.8$ and a dropout rate of $0.2$ during the post-folding phase. 
The pre-folding phase consists of $10$ epochs and the post-folding phase consists of $25$ epochs.

For latency measurements we used a batch size of $16$. We tested inference latency using Tensorflow’s built-in benchmarking tool (\verb+tf.test.Benchmark+) on NVIDIA Titan X Pascal GPU, averaged on 100 iterations with 10 burn-in iterations.

Our code is available at \url{https://github.com/LayerFolding/Layer-Folding}.


\end{document}